\documentclass{article}

\usepackage{PRIMEarxiv}

\usepackage[utf8]{inputenc} 
\usepackage[T1]{fontenc}    
\usepackage{hyperref}       
\usepackage{url}            
\usepackage{booktabs}       
\usepackage{amsfonts}       
\usepackage{nicefrac}       
\usepackage{microtype}      
\usepackage{lipsum}
\usepackage{fancyhdr}       
\usepackage{graphicx}       
\graphicspath{{media/}}     

\usepackage{lineno}
\usepackage{graphicx}
\usepackage{natbib}
\usepackage{hyperref}
\usepackage{amsmath} 
\usepackage{amssymb}
\usepackage{xcolor}
\usepackage{subcaption}
\usepackage{tabularx}
\usepackage{multirow}
\usepackage{algorithm}
\usepackage{algpseudocode}
\usepackage{comment}
\usepackage{float}
\usepackage{placeins}
\usepackage{booktabs}
\usepackage{makecell}
\usepackage{array}

\title{Kernel-Adaptive PI-ELMs for Forward and Inverse Problems in PDEs with Sharp Gradients
}

\author{
	Vikas Dwivedi\thanks{Corresponding Author} \\
	CREATIS Biomedical Imaging Laboratory \\
	INSA, CNRS UMR 5220, Inserm, Universit´e Lyon 1 \\
	Lyon 69621\\
	\texttt{vikas.dwivedi@creatis.insa-lyon.fr} \\
	\And
	Balaji Srinivasan \\
	Wadhwani School of Data Science \& AI \\
	Indian Institute of Technology Madras \\
	Chennai 600036\\
	\texttt{sbalaji@iitm.ac.in} \\
	\And
	Monica Sigovan \\
	CREATIS Biomedical Imaging Laboratory \\
	INSA, CNRS UMR 5220, Inserm, Universit´e Lyon 1 \\
	Lyon 69621\\
	\texttt{monica.sigovan@insa-lyon.fr} \\
	\And
	Bruno Sixou \\
	CREATIS Biomedical Imaging Laboratory \\
	INSA, CNRS UMR 5220, Inserm, Universit´e Lyon 1 \\
	Lyon 69621\\
	\texttt{bruno.sixou@insa-lyon.fr} \\
}

\begin{document}
\maketitle

\begin{abstract}
Physics-informed machine learning frameworks such as Physics-Informed Neural Networks (PINNs) and Physics-Informed Extreme Learning Machines (PI-ELMs) have shown great promise for solving partial differential equations (PDEs) but struggle with localized sharp gradients and singularly perturbed regimes, PINNs due to spectral bias and PI-ELMs due to their single-shot, non-adaptive formulation. We propose the Kernel-Adaptive Physics-Informed Extreme Learning Machine (KAPI-ELM), which performs Bayesian optimization over a low-dimensional, physically interpretable hyperparameter space governing the distribution of Radial Basis Function (RBF) centers and widths. This converts high-dimensional weight optimization into a low-dimensional distributional search, enabling targeted kernel refinement in regions with sharp gradients while also improving baseline solutions in smooth-flow regimes by tuning RBF supports. KAPI-ELM is validated on benchmark forward and inverse problems (1D convection--diffusion and 2D Poisson) involving PDEs with sharp gradients. It accurately resolves steep layers, improves smooth-solution fidelity, and recovers physical parameters robustly, matching or surpassing advanced methods such as the extended Theory of Functional Connections (X-TFC) with nearly an order of magnitude fewer tunable parameters. An extension to nonlinear problems is demonstrated by a curriculum-based solution of the steady Navier–Stokes equations via successive linearizations, yielding stable solutions for benchmark lid-driven cavity flow up to $Re=100$. These results indicate that KAPI-ELM provides an efficient and unified approach for forward and inverse PDEs, particularly in challenging sharp-gradient regimes.
\end{abstract}


\section{Introduction}\label{Sec:Introduction}
Physics-informed learning frameworks such as Physics-Informed Neural Networks (PINNs) \cite{RAISSI2019686} and Physics-Informed Extreme Learning Machines (PI-ELMs) \cite{DWIVEDI202096} have emerged as powerful tools for solving forward and inverse problems governed by partial differential equations (PDEs). Despite their growing success, both frameworks face limitations when applied to singularly perturbed PDEs characterized by sharp gradients and boundary layers. PINNs often suffer from spectral bias \cite{pmlr-v97-rahaman19a,WANG2022110768}, which hinders the representation of high-frequency components, while PI-ELMs rely on fixed, non-adaptive kernel distributions, limiting their ability to resolve localized features.

Several enhanced PINN formulations---including NTK-guided \cite{WANG2022110768}, self-adaptive \cite{MCCLENNY2023111722}, variable-scaling \cite{KO2025113860}, and curriculum-learning-based approaches \cite{NEURIPS2021_df438e52}---partially alleviate spectral bias and improve training dynamics. However, these methods remain computationally intensive and often unstable in singularly perturbed regimes. Conversely, lightweight alternatives such as PI-ELMs \cite{DWIVEDI202096,DONG2021114129} and the Extended Theory of Functional Connections (XTFC) \cite{SCHIASSI2021334,CALABRO2021114188} offer rapid convergence and analytical boundary satisfaction. Yet, their fixed input-layer design makes them sensitive to kernel placement and hyperparameters, restricting scalability and limiting their capability for inverse inference. This contrast highlights a persistent methodological gap between expressive but computationally expensive deep models and efficient but rigid shallow networks.

This work seeks to bridge that gap by asking:  
\begin{quote}
	\textit{\textbf{Can we enhance PI-ELMs with adaptive capability while preserving their computational simplicity?}}
\end{quote}

We propose the Kernel-Adaptive Physics-Informed Extreme Learning Machine (KAPI-ELM), a distributionally optimized extension of PI-ELM designed to capture sharp-gradients in both forward and inverse PDEs. Rather than tuning a large number of Radial Basis Function (RBF) centers and widths individually, KAPI-ELM introduces a low-dimensional, physically interpretable set of hyperparameters governing their statistical distributions. These hyperparameters are optimized via Bayesian optimization, transforming the high-dimensional weight search of conventional PI-ELMs into a compact, low-dimensional distributional search. The output-layer coefficients are still obtained through a single least-squares step, preserving the hallmark efficiency of PI-ELMs while enabling adaptive representation of localized physics.

The proposed framework is validated on forward and inverse PDEs with sharp gradients, including benchmark 1D convection--diffusion equations with single and twin boundary layers and a 2D Poisson equation with a sharp localized source. KAPI-ELM accurately resolves steep gradients and recovers unknown parameters robustly, matching or even outperforming XTFC and PINN counterparts while requiring nearly an order of magnitude fewer tunable parameters. We further develop a curriculum learning-driven KAPI-ELM to solve non-linear PDEs by successive linearizations, giving accurate solution of the benchmark lid-driven cavity up to a $Re=100$. The method thus unifies efficiency, interpretability, and adaptivity within a single, physics-informed learning framework.

\paragraph{\textbf{Organization}}
The remainder of this paper is organized as follows.
Section~\ref{Sec:Prelim} briefly recaps the Physics-Informed Extreme Learning Machine (PI--ELM) framework and its formulation using fixed Gaussian radial basis functions.
Section~\ref{Sec:KAPI} introduces the proposed Kernel-Adaptive Physics-Informed Extreme Learning Machine (KAPI--ELM), detailing the distributional parameterization of RBF centers and widths, the nested (inner–outer) optimization strategy, and its extension to inverse problems.
Section~\ref{Sec:Results} presents comprehensive numerical experiments on forward and inverse problems, including singularly perturbed 1D convection--diffusion with single and twin boundary layers, a 2D Poisson equation with a sharply localized source, and a nonlinear Navier--Stokes lid-driven cavity problem solved through curriculum learning (Section~\ref{sec:lid-driven}).
Section~\ref{sec:limitations} discusses the main limitations of the current framework.
Finally, Section~\ref{Sec:Conclusion} concludes the paper with key findings and perspectives for future research.


\section{Preliminaries: Physics-Informed Extreme Learning Machine (PI-ELM)}
\label{Sec:Prelim}

Consider a linear differential operator $\mathcal{L}_{\nu}(\cdot)$ defined over a domain $\Omega \subset \mathbb{R}$ with boundary $\partial \Omega$, parameterized by $\nu$. The PI-ELM~\cite{DWIVEDI202096} seeks an approximate solution $u(x)$ satisfying
\begin{equation}
	\mathcal{L}_{\nu}(u) = R(x), \quad x \in \Omega, 
	\qquad 
	u(x) = g(x), \quad x \in \partial \Omega,
	\label{eq:pde}
\end{equation}
where $R(x)$ is the source term and $g(x)$ the boundary condition.

The solution is represented as a global weighted sum of $N^*$ fixed Gaussian basis functions,
\begin{equation}
	\hat{u}(x) = \sum_{i=1}^{N^*} c_i\, 
	\exp\!\left[-\frac{(x-\alpha_i^*)^2}{2\sigma_i^2}\right],
	\label{eq:pielm_hypothesis}
\end{equation}
where $\alpha_i^*$ and $\sigma_i$ denote the RBF centers and widths, randomly sampled from prescribed distributions, and $\mathbf{c}=[c_1,\dots,c_{N^*}]^\top$ are the output weights determined analytically.

For compactness, each basis function can equivalently be expressed in a neural form as
\begin{equation}
	G_i(x;m_i,b_i)
	= \exp\!\big[-(m_i x + b_i)^2\big],
	\label{eq:pielm_mx_b_form}
\end{equation}
where $m_i = \tfrac{1}{\sqrt{2}\sigma_i}$ and $b_i = -\tfrac{\alpha_i^*}{\sqrt{2}\sigma_i}$.

By enforcing the PDE residual and boundary conditions at $N_c$ interior and $N_{bc}$ boundary collocation points, one obtains the linear system
\begin{equation}
	H\mathbf{c}=\mathbf{r},
	\label{eq:pielm_linear}
\end{equation}
where each row of \( H \in \mathbb{R}^{(N_c + N_{bc}) \times N^*} \)  corresponds to the evaluation of $\mathcal{L}_{\nu}$ or the boundary operator at a specific point, and \( \mathbf{r} \in \mathbb{R}^{(N_c + N_{bc})} \) collects the target values. The optimal coefficients are obtained using the Moore--Penrose pseudoinverse as
\begin{equation}
	\mathbf{c} = H^{\dagger}\, \mathbf{r},
	\label{eq:pielm_solution}
\end{equation}
which yields the minimum-norm least-squares solution to $\|H\mathbf{c} - \mathbf{r}\|_2^2$.

Unlike PINNs \cite{RAISSI2019686}, which iteratively update all network parameters via backpropagation and use deep architectures to encode physics constraints, PI-ELM with a fixed input layer offers a single-shot least-squares solution without backpropagation, enabling extremely fast training. However, its performance strongly depends on the randomly initialized RBF parameters $(\alpha_i^*, \sigma_i)$, which remain fixed during optimization. This non-adaptive nature limits its accuracy in resolving localized features or sharp-gradient regions—a key motivation for the kernel-adaptive extension proposed next.

\FloatBarrier
\section{Kernel-Adaptive Physics-Informed ELM (KAPI-ELM)}
\label{Sec:KAPI}
\begin{figure}[h]
	\centering
	\includegraphics[width=0.8\linewidth]{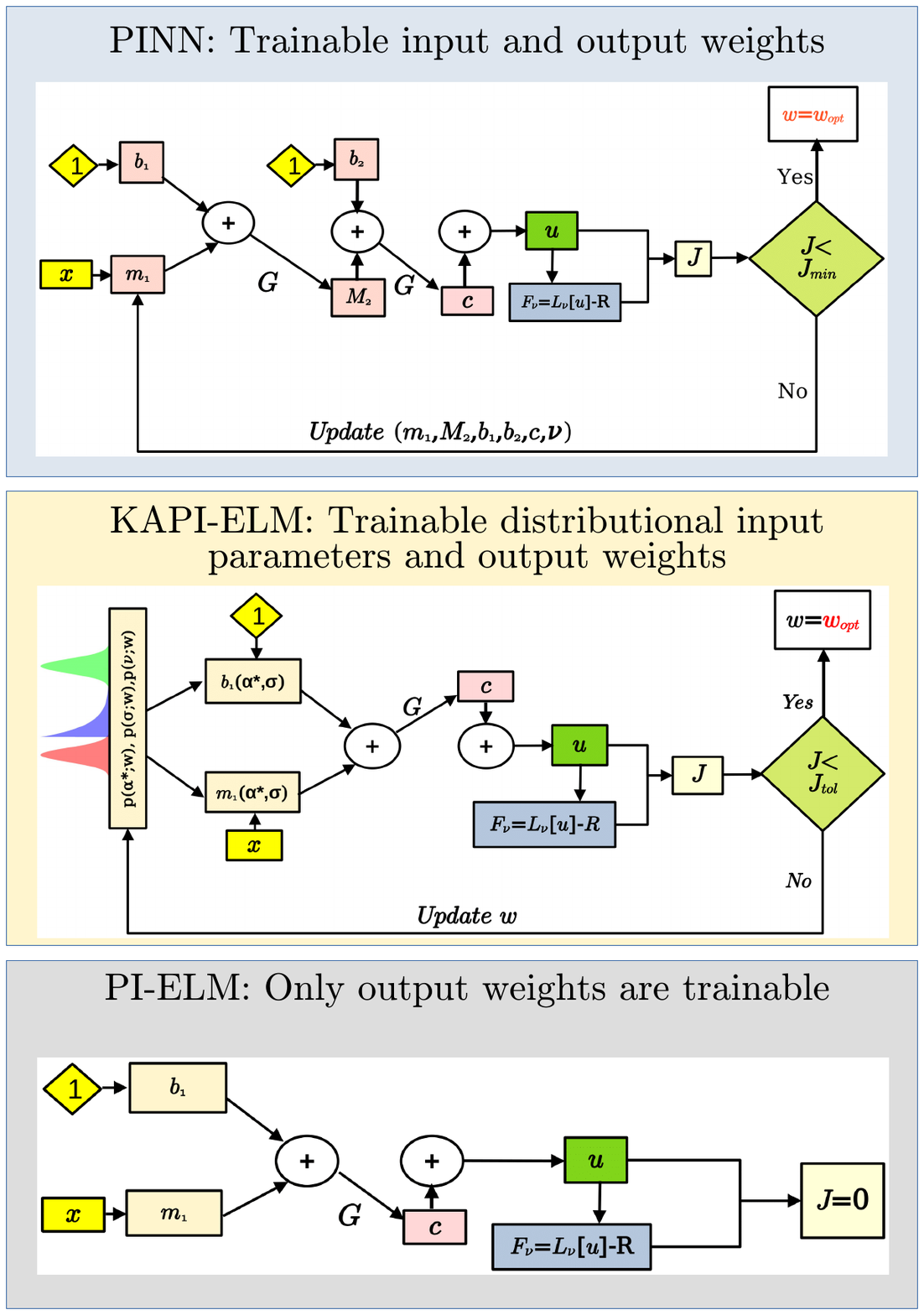}
	\caption{\textit{Conceptual continuum of physics-informed learners.}
		PINNs train all input and output weights via backpropagation, offering flexibility but at a high computational cost.
		PI-ELMs fix random input weights and determine only the output weights in a single shot least square solve.
		KAPI-ELM bridges these extremes through \textbf{distributional optimization} of input parameters, achieving adaptive kernel learning without backpropagation while retaining the \textbf{analytic least-squares solution} for output weights.}
	\label{fig:pinn_pielm_kapielm}
\end{figure}

The proposed Kernel-Adaptive Physics-Informed Extreme Learning Machine (KAPI-ELM) extends the PI-ELM philosophy by introducing trainable kernel distributions instead of direct weight optimization. Figure~\ref{fig:pinn_pielm_kapielm} contrasts KAPI-ELM with two limiting paradigms—Physics-Informed Neural Networks (PINNs), which train all weights via iterative backpropagation, and Physics-Informed Extreme Learning Machines (PI-ELMs), which fix random input weights and solve the output layer analytically in single shot. KAPI-ELM lies between these extremes, leveraging \textit{distributional optimization} of input kernels to achieve adaptive representation while preserving the analytic least-squares efficiency of ELMs.

\subsection{Model Problem}

For simplicity, we illustrate the proposed method using the one-dimensional steady convection–diffusion equation,
\begin{equation}
	-\nu\,u_{xx} + u_x = 0, \qquad x \in [0,1],
	\label{eq:conv_diff}
\end{equation}
subject to Dirichlet boundary conditions
\begin{equation}
	u(0)=0, \qquad u(1)=1.
	\label{eq:bc}
\end{equation}
The exact analytical solution is
\begin{equation}
	u(x) = \frac{e^{x/\nu}-1}{e^{1/\nu}-1},
\end{equation}
which exhibits a thin boundary layer near $x=1$ when the diffusion coefficient $\nu$ is small.  
As $\nu \to 0$, the problem becomes singularly perturbed, and conventional fixed-weight PI-ELMs computationally heavy to capture the steep gradient in the boundary layer region. The proposed Kernel-Adaptive PI-ELM (KAPI-ELM) overcomes this limitation by adaptively redistributing the RBF kernels based on a small set of tunable distributional hyperparameters.

\subsection{Forward Problem Setup}
\paragraph{Trial Solution}
Following the standard PI-ELM formulation (see Section~\ref{Sec:Prelim}), the approximate solution is expressed as a global weighted sum of $N^*$ Gaussian radial basis functions where the RBF centers $\alpha_i^*$ and widths $\sigma_i$ are \emph{not fixed}, but treated as random variables drawn from parameterized distributions. The goal is to determine the parameters of these distributions so that the resulting basis adapts to the underlying physics of the PDE.

\paragraph{RBF Centers Distribution}
To ensure both global coverage and local adaptivity, we define a two-component mixture model for the RBF centers:
\begin{equation}
	p(\alpha^*;w) = \pi_{\mathrm{base}}\,\mathcal{U}(0,1)
	+ \pi_{\mathrm{adap}}\,\mathcal{N}(\mu,\tau^2),
	\label{eq:kapi_mixture}
\end{equation}
where $\pi_{\mathrm{base}}$ and $\pi_{\mathrm{adap}}$ are mixture weights satisfying $\pi_{\mathrm{base}}+\pi_{\mathrm{adap}}=1$.  
The baseline component provides quasi-uniform coverage across the domain, implemented via evenly spaced centers approximating a uniform distribution, while the adaptive component concentrates RBFs near regions of sharp gradients (e.g., boundary layers). The fraction of RBFs allocated to the adaptive component is denoted by $f=\pi_{\mathrm{adap}}/\pi_{\mathrm{base}}$. The adaptive component in \eqref{eq:kapi_mixture} is parameterized by its mean $\mu$ and standard deviation $\tau$. The mean $\mu$ determines the region of interest (e.g., near $x=1$ for an outflow boundary layer), while $\tau$ controls the spatial spread of adaptive RBFs. This formulation allows both collocation points and RBF centers to be sampled from the same distribution, with adaptive centers co-located with collocation points in sharp-gradient regions.  

\paragraph{RBF Widths Distribution}
The RBF widths combine a fixed baseline scale with an adaptive refinement driven by the PDE stiffness.  
For the uniformly distributed (baseline) RBFs, constant widths are assigned as
\[
\sigma = \sigma_F = \kappa/N_{\mathrm{unif}},
\]
where $\kappa$ is an overlap factor ensuring smoothness and numerical stability.  

For the adaptive RBFs, widths are stochastically generated from a bounded inverse-scale distribution,
\[
\sigma_k = \min\!\left(\left|\frac{1}{\sqrt{2}\,\xi_k}\right|,\,\sigma_F\right),
\qquad
\xi_k \sim \mathcal{U}\!\left[-\tfrac{\zeta_k}{2},\,\tfrac{\zeta_k}{2}\right],
\]
where $\xi_k$ is a random inverse-scale parameter controlling local sharpness.  
The upper bound $\zeta_k$ is dynamically linked to the diffusion coefficient $\nu$ as
\[
\zeta_k = \frac{1}{\sqrt{2}\,\sigma_F}\,\nu^{-(1+\lambda_k)},
\]
with $\lambda_k$ acting as a tunable decay factor that controls how sensitively the kernel widths respond to stiffness.  
As $\nu$ decreases, $\zeta_k$ grows exponentially, resulting in narrower kernels and enhanced spatial resolution near boundary layers, while the cap at $\sigma_F$ prevents numerical instability in smoother regions.

\paragraph{Distributional Hyperparameters}
The mixture fraction $f$, mean $\mu$, spread $\tau$, and width-decay exponent $\lambda$ together define the low-dimensional hyperparameter vector
\begin{equation}
	w = [f, \mu, \tau, \lambda].
	\label{eq:w_params}
\end{equation}
Hence, instead of directly optimizing a large number of RBF parameters $\{\alpha_i^*,\sigma_i\}_{i=1}^{N^*}$, KAPI-ELM optimizes only the four hyperparameters in $w$ that govern their statistical distribution.  
This significantly reduces the optimization space while preserving interpretability.

\paragraph{Optimization Strategy}
The training of KAPI-ELM proceeds in two nested stages of optimization. 
For a given set of distributional hyperparameters $\boldsymbol{w}$, the inner problem determines the optimal RBF output weights $\mathbf{c}$ by minimizing the physics-informed residual loss:
\begin{equation}
	\mathbf{c}^{*}(\boldsymbol{w}) 
	= \arg \min_{\mathbf{c}} 
	\|H(\boldsymbol{\alpha}^{*},\boldsymbol{\sigma}\mid\boldsymbol{w})\mathbf{c}-\mathbf{r}\|_2^2,
	\label{eq:inner_opt}
\end{equation}
where $H(\boldsymbol{\alpha}^{*},\boldsymbol{\sigma}\mid\boldsymbol{w})$ is the design matrix constructed using RBF kernels sampled from the parameterized distributions $p(\alpha^*;\boldsymbol{w})$ and $p(\sigma;\boldsymbol{w},\nu)$, and $\mathbf{r}$ contains the PDE residuals and boundary data.
The inner optimization admits the closed-form least-squares solution
\begin{equation}
	\mathbf{c}^{*}(\boldsymbol{w}) = 
	H^{\dagger}\, \mathbf{r},
\end{equation}
which is the minimum-norm solution of the linear system in the least-squares sense.

Substituting this analytic solution into the outer objective yields the physics-informed loss 
evaluated on an independent validation set,
\begin{equation}
	J(\boldsymbol{w}) 
	= \|H_{\mathrm{val}}(\boldsymbol{\alpha}^{*},\boldsymbol{\sigma}\mid\boldsymbol{w})
	\,\mathbf{c}^{*}(\boldsymbol{w}) - \mathbf{r}_{\mathrm{val}}\|_{\infty},
	\label{eq:outer_loss}
\end{equation}
where $H_{\mathrm{val}}$ and $\mathbf{r}_{\mathrm{val}}$ denote the design matrix and target vector evaluated at unseen collocation points.

The optimal hyperparameters are then obtained as
\begin{equation}
	\boldsymbol{w}^{*} 
	= \arg \min_{\boldsymbol{w} \in [\boldsymbol{w}_{\mathrm{lb}},\,\boldsymbol{w}_{\mathrm{ub}}]}
	J(\boldsymbol{w}),
	\label{eq:outer_opt}
\end{equation}
where $[\boldsymbol{w}_{\mathrm{lb}},\,\boldsymbol{w}_{\mathrm{ub}}]$ denotes the user-defined, physics-informed bounds on the distributional parameters. Equation~\eqref{eq:outer_opt} is solved via Bayesian optimization with a Gaussian-process surrogate and expected-improvement acquisition criterion.  
This nested optimization couples the analytical efficiency of PI-ELM’s least-squares solution with a physics-informed distributional search over $\boldsymbol{w}$, enabling adaptive kernel placement without backpropagation or gradient computation.

\subsection{Inverse Problem Setup}
We identify the diffusion coefficient $\nu$ from sparse measurements while enforcing the PDE. Let $\boldsymbol{w}$ denote the distributional hyperparameters and define the stacked outer variable
\begin{equation}
	\vartheta \;=\; [\,\boldsymbol{w};\,\nu\,].
\end{equation}
For fixed $\vartheta$, the inner problem determines the RBF output weights by minimizing the physics residual (PDE $+$ boundary terms) on a training collocation set:
\begin{equation}
	\mathbf{c}^{*}(\vartheta)\;=\;\arg\min_{\mathbf{c}}\;\big\|H(\vartheta)\,\mathbf{c}-\mathbf{r}\big\|_2^2,
	\label{eq:inv_inner}
\end{equation}
which admits the closed-form solution $\mathbf{c}^{*}(\vartheta)=H(\vartheta)^{\dagger}\,\mathbf{r}$.

Let $\Phi_{\mathrm{obs}}(\vartheta)$ be the RBF evaluation matrix at measurement locations $\{x_j\}_{j=1}^{N_m}$, so the predicted observations are
\begin{equation}
	\hat{\mathbf{u}}_{\mathrm{obs}}(\vartheta)\;=\;\Phi_{\mathrm{obs}}(\vartheta)\,\mathbf{c}^{*}(\vartheta).
\end{equation}
The outer objective penalizes both data misfit and PDE misfit (evaluated on an independent validation set):
\begin{equation}
	J_{\mathrm{inv}}(\vartheta)\;=\;\big\|\hat{\mathbf{u}}_{\mathrm{obs}}(\vartheta)-\mathbf{u}_{\mathrm{obs}}\big\|_2^2
	\;+\;\lambda_{\mathrm{pde}}\;\big\|H_{\mathrm{val}}(\vartheta)\,\mathbf{c}^{*}(\vartheta)-\mathbf{r}_{\mathrm{val}}\big\|_2^2,
	\label{eq:inv_outer_loss}
\end{equation}
where $\lambda_{\mathrm{pde}}>0$ balances data fidelity and physics consistency. The inverse problem is
\begin{equation}
	\vartheta^{*}
	\;=\;
	\arg\min_{\vartheta \in [\vartheta_{\mathrm{lb}},\,\vartheta_{\mathrm{ub}}]}
	J_{\mathrm{inv}}(\vartheta),
	\label{eq:inv_outer_opt}
\end{equation}
where $[\vartheta_{\mathrm{lb}},\,\vartheta_{\mathrm{ub}}]$ defines the user-specified, physics-informed bounds on both the distributional parameters $\boldsymbol{w}$ and the physical coefficient $\nu$.  
This outer optimization is performed via Bayesian optimization with a Gaussian-process surrogate and expected improvement. This nests analytic least-squares estimation of $\mathbf{c}$ within a distributional search over $\boldsymbol{w}$ and $\nu$, ensuring parameter recovery is simultaneously constrained by data and PDE physics.

Although we have described KAPI-ELM for 1D steady model problem with one sharp jump for clarity, extending it to 2D and multiple jump problems is straightforward.
\FloatBarrier
\section{Results and Discussion}
\label{Sec:Results}
We evaluate the performance of the proposed KAPI-ELM framework on representative forward and inverse problems involving localized sharp gradients and stiffness. 
All computations are performed in MATLAB~R2022b on a 12th-Gen Intel(R) Core(TM) i7-12700H (2.30 GHz, 16 GB RAM) laptop. 

\subsection{Forward Problems}
\subsubsection{Test Cases}
We consider three benchmark forward problems:  
(i) a singularly perturbed 1D convection–diffusion equation with a single boundary layer~\cite{DEFLORIO2024115396},  
(ii) a modified convection–diffusion equation exhibiting symmetric twin layers~\cite{KUMAR20152081}, and  
(iii) a 2D Poisson equation with a sharply localized Gaussian source~\cite{Rodr_guez_Lara_2025}.  
Table~\ref{Tab:ForwardCases} summarizes the governing equations, boundary conditions, and reference solutions. As the diffusion parameter~$\nu$ decreases, each case develops increasingly localized gradients, making them suitable for evaluating the stiffness-resolving capability of KAPI-ELM across both 1D and 2D regimes.

\begin{table}[h!]
	\centering
	\caption{Summary of forward-problem test cases.}
	\label{Tab:ForwardCases}
	\renewcommand{\arraystretch}{1.2}
	\setlength{\tabcolsep}{4pt}
	\begin{tabular}{|p{2cm}|p{3.5cm}|p{3.4cm}|p{3cm}|}
		\hline
		\textbf{Test Case} & \textbf{Governing Equation} & \textbf{BCs} & \textbf{Ground Truth} \\
		\hline
		Single BL &
		$u_x - \nu u_{xx} = 0$ &
		$u(0)=0,\; u(1)=1$ &
		$\dfrac{e^{x/\nu}-1}{e^{1/\nu}-1}$ \\
		\hline
		Twin BL &
		$2(2x-1)u_x - \nu u_{xx} + 4u = 0$ &
		$u(0)=1,\; u(1)=1$ &
		$e^{-2x(1-x)/\nu}$ \\
		\hline
		2D Poisson &
		$\nabla^2 u =\tfrac{1}{2\pi\nu^2}\exp[-((x-0.5)^2+(y-0.5)^2)/(2\nu^2)]$ &
		$u|_{\partial\Omega}=0$ &
		Finite Difference Solution \\
		\hline
	\end{tabular}
\end{table}

\subsubsection{Boundary Layer Problems}
Both single and twin boundary-layer (BL) problems are solved under identical numerical settings: 
$N_{unif}=1500$, $N_{bc}=2$, and $\sigma_F=0.0133$. 
Dirichlet boundary conditions are imposed in all cases. 
The adaptive kernel distribution is parameterized by tunable variables $\boldsymbol{w}$, 
optimized via Bayesian optimization with 200 evaluations of the expected-improvement-plus acquisition.

For the single BL problem, $\boldsymbol{w}=[f,\mu,\tau,\lambda]$, 
while the twin BL employs two symmetric components 
$\boldsymbol{w}=[f_1,\mu_1,\tau_1,\lambda_1,f_2,\mu_2,\tau_2,\lambda_2]$
to capture layers near both boundaries. 
For $\nu=10^{-2}$ and $10^{-3}$, the single BL uses bounds 
$\boldsymbol{w}_{\mathrm{lb}}=[0.495,0.90,0.10,0.50]$, 
$\boldsymbol{w}_{\mathrm{ub}}=[0.505,0.99,0.40,1.00]$, 
tightened for $\nu=10^{-4}$ to 
$\boldsymbol{w}_{\mathrm{lb}}=[0.495,0.99,0.01,0.50]$, 
$\boldsymbol{w}_{\mathrm{ub}}=[0.505,0.999,0.10,1.00]$. 
The twin BL problem uses uniform bounds across all $\nu$ values:
\[
\begin{bmatrix}
	\boldsymbol{w}_{\mathrm{lb}} \\[3pt]
	\boldsymbol{w}_{\mathrm{ub}}
\end{bmatrix}
=
\begin{bmatrix}
	0.495 & 0.495 & 0.001 & 0.99  & 0.01 & 0.01 & 0.5 & 0.5 \\
	0.505 & 0.505 & 0.010 & 0.999 & 0.10 & 0.10 & 1.0 & 1.0
\end{bmatrix}.
\]

\begin{figure}[h]
	\centering
	\includegraphics[width=\linewidth]{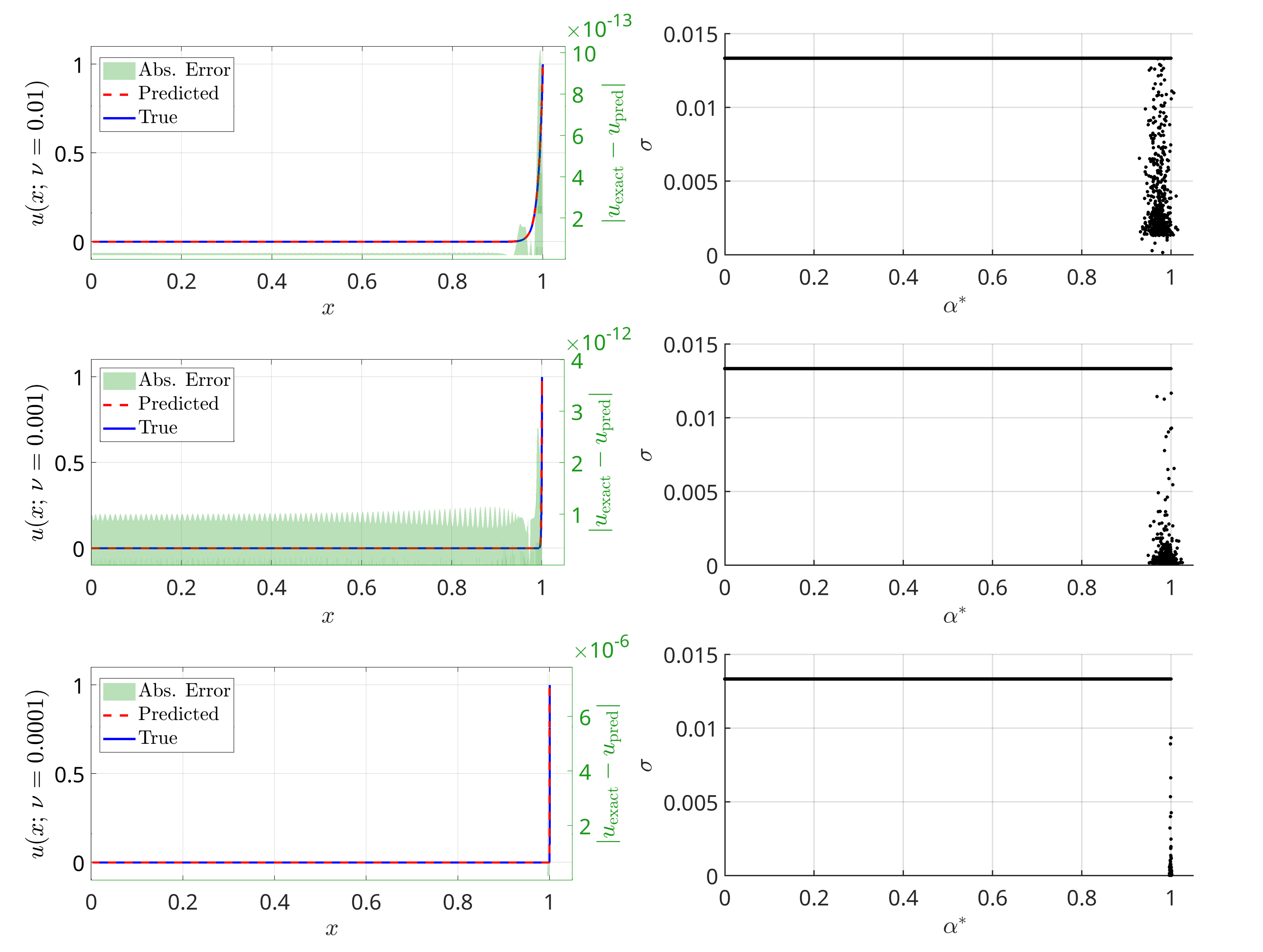}
	\caption{
		Predicted and exact solutions (left) and corresponding RBF distributions (right)
		for the single boundary-layer problem across stiffness levels. 
		As $\nu$ decreases, the adaptive component automatically concentrates RBF centers 
		near the outflow boundary ($x=1$), capturing the thin layer without manual refinement.
	}
	\label{Fig:SingleBL}
\end{figure}

\begin{figure}[h]
	\centering
	\includegraphics[width=\linewidth]{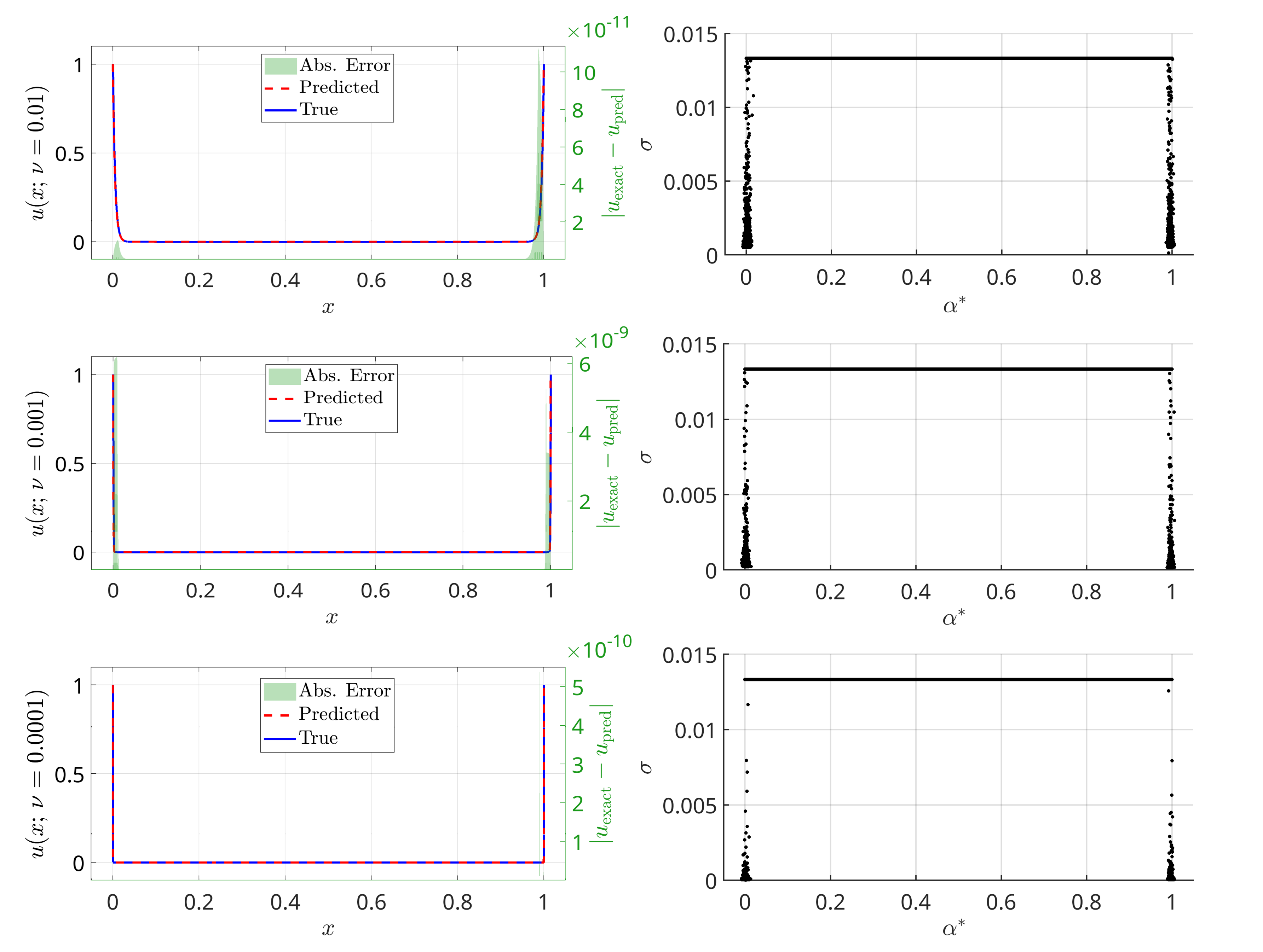}
	\caption{
		Predicted versus exact solutions (left) and corresponding RBF distributions (right)
		for the twin boundary-layer problem. 
		The two adaptive Gaussian components automatically localize near $x=0$ and $x=1$, 
		resolving both layers symmetrically as $\nu$ decreases.
	}
	\label{Fig:TwinBL}
\end{figure}

The optimized hyperparameters and wall-clock times for both cases are summarized in 
Tables~\ref{Tab:SingleBL} and~\ref{Tab:TwinBL}, while the corresponding solutions 
and RBF distributions are shown in Figures~\ref{Fig:SingleBL} and~\ref{Fig:TwinBL}.

\begin{table}[h!]
	\centering
	\caption{Optimized hyperparameters and performance for the single boundary-layer problem.}
	\label{Tab:SingleBL}
	\renewcommand{\arraystretch}{1.1}
	\setlength{\tabcolsep}{5pt}
	\begin{tabular}{|c|c|c|c|c|c|c|}
		\hline
		$\boldsymbol{\nu}$ & $f$ & $\mu$ & $\tau$ & $\lambda$ & \textbf{Time (s)} & $\log_{10}(J_{\min})$ \\
		\hline
		$10^{-2}$ & 0.4972 & 0.9722 & 0.3970 & 0.505 & 4.09 & $-8.34$ \\
		$10^{-3}$ & 0.5050 & 0.9872 & 0.3539 & 0.756 & 2.42 & $-9.65$ \\
		$10^{-4}$ & 0.5043 & 0.9990 & 0.0248 & 0.705 & 6.53 & $-8.51$ \\
		\hline
	\end{tabular}
\end{table}

\begin{table}[h!]
	\centering
	\caption{Optimized hyperparameters and performance for the twin boundary-layer problem.}
	\label{Tab:TwinBL}
	\renewcommand{\arraystretch}{1.1}
	\setlength{\tabcolsep}{3pt}
	\begin{tabular}{|c|cccc|cccc|c|c|}
		\hline
		$\boldsymbol{\nu}$ &
		$f_1$ & $\mu_1$ & $\tau_1$ & $\lambda_1$ &
		$f_2$ & $\mu_2$ & $\tau_2$ & $\lambda_2$ &
		\textbf{Time (s)} & $\log_{10}(J_{\min})$ \\
		\hline
		$10^{-2}$ & 0.5047 & 0.5013 & 0.0031 & 0.995 & 0.0999 & 0.0916 & 0.9929 & 0.978 & 17.8 & $-8.56$ \\
		$10^{-3}$ & 0.5001 & 0.5020 & 0.0017 & 0.997 & 0.0878 & 0.0850 & 0.615 & 0.715 & 44.9 & $-8.25$ \\
		$10^{-4}$ & 0.4953 & 0.4957 & 0.0011 & 0.997 & 0.0846 & 0.0793 & 0.734 & 0.717 & 111.0 & $-8.14$ \\
		\hline
	\end{tabular}
\end{table}

Across all stiffness levels, KAPI-ELM maintains $\log_{10}(J_{\min})<-8$ 
and achieves convergence within seconds for the single BL and under two minutes for the twin BL cases. 
As $\nu$ decreases, $(\mu,\tau)$ (or $(\mu_{1,2},\tau_{1,2})$) consistently shift toward the 
boundary-layer regions and shrink in scale, evidencing automatic kernel localization 
consistent with the analytical layer thickness $\mathcal{O}(\nu)$. 
These results confirm that the distributional optimization in KAPI-ELM 
achieves robust, physics-consistent adaptivity without mesh refinement or gradient-based training.

\subsubsection{2D Poisson Equation with a Sharp Gaussian Source}

We next consider the 2D Poisson problem (Table~\ref{Tab:ForwardCases})
\[
\nabla^2 u = \frac{1}{2\pi\nu^2}\exp\!\Big[-\frac{(x-0.5)^2+(y-0.5)^2}{2\nu^2}\Big],
\qquad (x,y)\in[0,1]^2,\quad
u|_{\partial\Omega}=0,
\]
which features a strongly localized source near $(0.5,0.5)$ that induces steep gradients for small $\nu$.  
This test assesses the ability of KAPI-ELM to adaptively cluster kernels in 2D without mesh refinement.

For a two-dimensional domain $\Omega \subset \mathbb{R}^2$ with coordinates
$\boldsymbol{x}=(x,y)$, the PI--ELM approximation generalizes to a global
weighted sum of $N^*$ fixed Gaussian radial basis functions,
\begin{equation}
	\hat{u}(x,y)
	=
	\sum_{i=1}^{N^*} 
	c_i\,
	\exp\!\left(
	-\frac{
		(x-\mu_{x,i})^2
		+
		(y-\mu_{y,i})^2
	}{
		2\sigma_i^2
	}
	\right),
	\label{eq:pielm_hypothesis_2d}
\end{equation}
where $\boldsymbol{\alpha}_i^*=(\alpha_i^{(x)},\alpha_i^{(y)})$ are the RBF
centers, $\sigma_i$ are the corresponding widths, and 
$\boldsymbol{c}=[c_1,\dots,c_{N^*}]^\top$ denotes the output-layer weights
obtained analytically. 

We use fixed baseline parameters $N_{unif}=1600$, $\sigma_F=0.20$, and
$N_{bc}=1600$.  
The baseline configuration contributes $N_{rbf}^{\mathrm{base}}=400$ uniformly distributed RBFs.  
The adaptive component, parameterized by
$\boldsymbol{w}=[f,\mu_x,\mu_y,\tau,\lambda]$, 
is optimized via Bayesian optimization (200 evaluations, expected-improvement-plus acquisition).
Physics-based bounds are
$f\in[0.5,1.0]$, 
$\mu_x,\mu_y\in[0.2,0.8]$, 
$\tau\in[0.1,0.5]$, and 
$\lambda\in[0.5,0.9]$.

\begin{figure}[h]
	\centering
	\includegraphics[width=\linewidth]{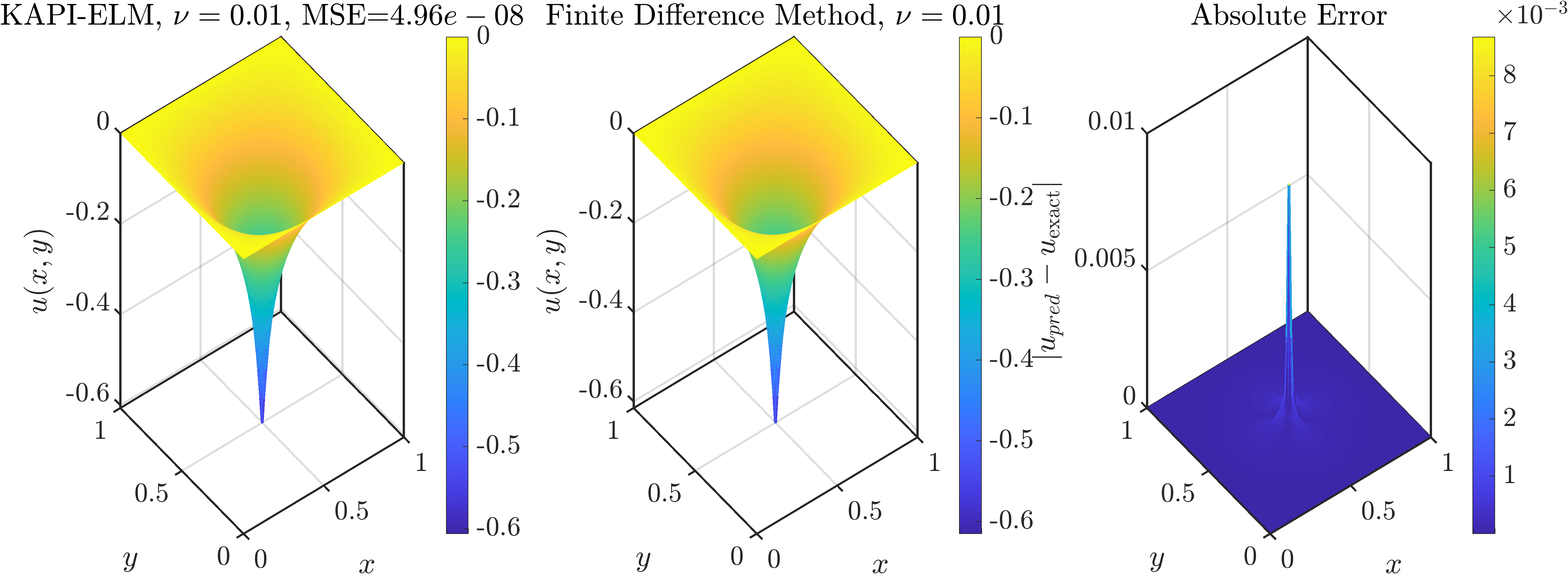}
	\caption{
		KAPI-ELM solution (left), finite-difference reference (center), and absolute error (right)
		for the 2D Poisson problem with $\nu=10^{-2}$.
		The adaptive component concentrates RBFs near $(0.5,0.5)$, sharply resolving the localized Gaussian source.
		Residuals remain below $10^{-4}$, and the mean-squared error is $\mathrm{MSE}=4.96\times10^{-8}$,
		demonstrating high accuracy and numerical stability.
	}
	\label{Fig:Poisson}
\end{figure}

For $\nu=10^{-2}$, the optimization converges in 119 evaluations and $61.3$\,s, yielding
$\boldsymbol{w}^*=[0.7723,\,0.4915,\,0.5068,\,0.1554,\,0.8869]$.
The learned configuration places $\mu_x,\mu_y$ near the domain center with a narrow spread $\tau\approx0.16$,
effectively concentrating RBFs around the localized source.
Table~\ref{Tab:Poisson} summarizes the optimized configuration and quantitative performance.
Figure~\ref{Fig:Poisson} compares the predicted field, the finite-difference reference, and the absolute error.

\begin{table}[h!]
	\centering
	\caption{Optimized hyperparameters and performance for the 2D Poisson problem ($\nu=10^{-2}$).}
	\label{Tab:Poisson}
	\renewcommand{\arraystretch}{1.1}
	\setlength{\tabcolsep}{5pt}
	\begin{tabular}{|c|c|c|c|c|c|c|c|c|c|}
		\hline
		$\boldsymbol{\nu}$ & $f$ & $\mu_x$ & $\mu_y$ & $\tau$ & $\lambda$ & $N_c$ & $N_{bc}$ & \textbf{Time (s)} & $L^2$\textbf{-error} \\
		\hline
		$10^{-2}$ & 0.7723 & 0.4915 & 0.5068 & 0.1554 & 0.8869 & 1600 & 1600 & 61.3 & $4.96\times10^{-8}$ \\
		\hline
	\end{tabular}
\end{table}

\begin{figure}[h]
	\centering
	\includegraphics[width=0.95\textwidth]{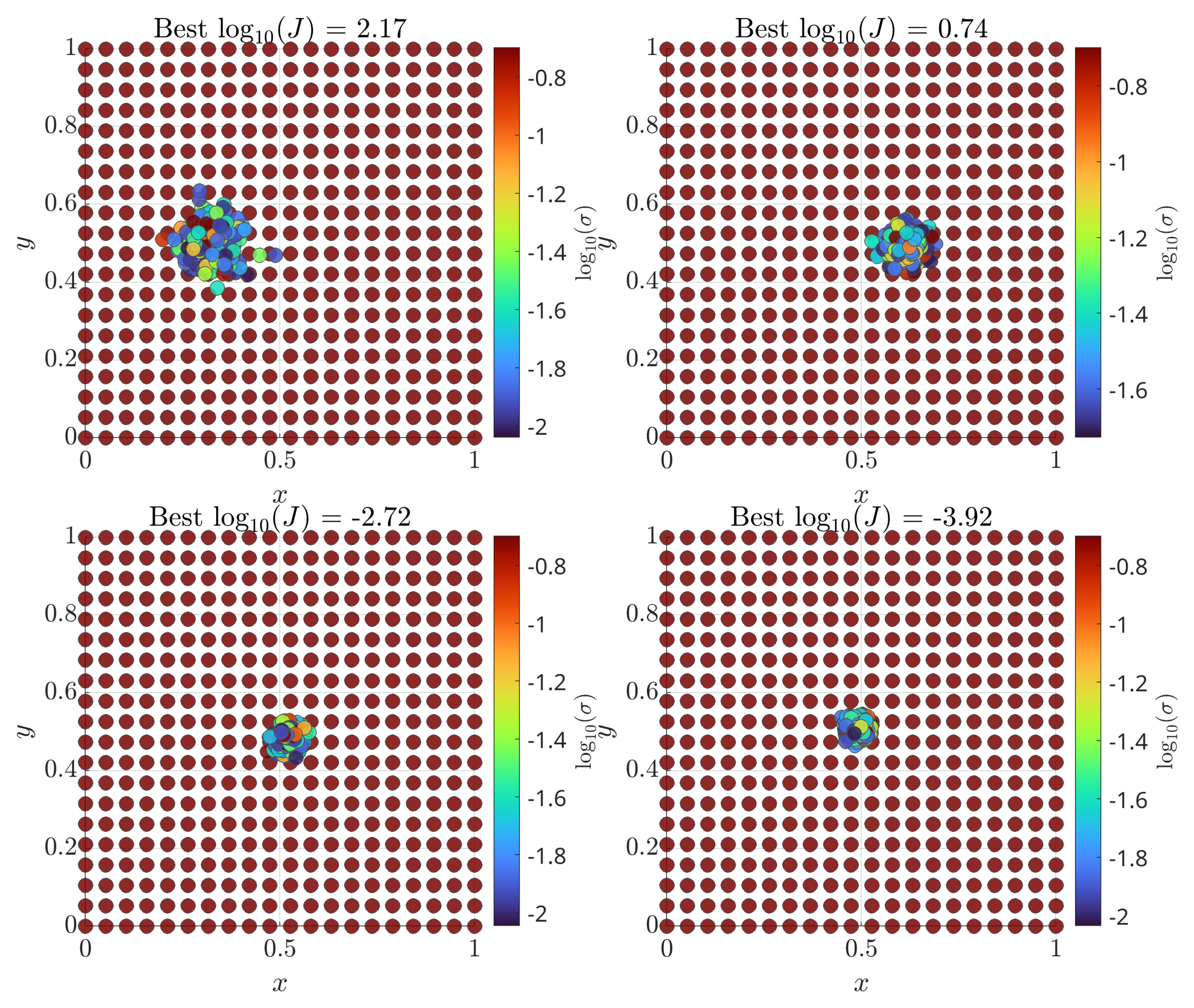}
	\caption{
		Evolution of the adaptive RBF kernel distribution during Bayesian optimization 
		for the 2D Poisson problem ($\nu = 10^{-2}$). 
		Each panel shows the RBF centers overlaid on the fixed baseline grid, colored by $\log_{10}(\sigma)$. 
		As optimization proceeds (top-left to bottom-right), the adaptive RBFs progressively 
		concentrate near the localized Gaussian source at $(x,y)\approx(0.5,0.5)$, while 
		kernel widths shrink to resolve the steep gradients. 
		The decreasing $\log_{10}(J)$ values indicate convergence toward the optimal configuration.
	}
	\label{Fig:RBF_Evolution}
\end{figure}

KAPI-ELM achieves an $L^2$ error below $10^{-7}$ and $\log_{10}(J_{\min})\approx-3.9$ (see Fig. \ref{Fig:RBF_Evolution}) within one minute of wall-clock time, without any mesh refinement or backpropagation.  The method accurately captures the steep, isotropic decay of the potential field around the source, confirming the robustness of the distributional optimization strategy in 2D, stiff PDE settings. The 2D Poisson case shows KAPI-ELM’s ability to generalize beyond 1D problems.

\subsection*{Remarks}
\begin{enumerate}
	\item \textit{Physics-informed hyperparameters.}
	The distributional hyperparameters $\boldsymbol{w}$ govern the adaptive RBF configuration (adaptive fraction, mean and standard deviation of the center distribution, and the width-decay exponent). Their physical interpretability enables tight, problem-aware search bounds. For example, in the single boundary-layer (BL) case we used $\boldsymbol{w}_{\mathrm{lb}}=[0.495,\,0.99,\,0.01,\,0.50]$ and $\boldsymbol{w}_{\mathrm{ub}}=[0.505,\,0.999,\,0.10,\,1.00]$ at $\nu=10^{-4}$, which guided the optimizer to concentrate centers near $x=1$ with small spread (e.g., $\mu\approx 0.999$, $\tau\approx 0.025$), consistent with the layer thickness $\mathcal{O}(\nu)$. For a related application of this concept in physics-informed image segmentation, we refer readers to \cite{Dwivedi2024}.
	
	\item \textit{Dimensionality reduction.}
	KAPI-ELM encodes a high-dimensional kernel configuration (e.g., $N_r=1500$ RBF centers and widths in the single boundary-layer problem) into a compact, interpretable hyperparameter vector $\boldsymbol{w}$ of dimension $N_w\ll N_r$ (e.g., $N_w=4$ for single BL, $N_w=8$ for twin BL). This mapping reduces the search space by several orders of magnitude while retaining adaptive expressiveness through distributional control. Figure~\ref{Fig:BL_Interpretation} illustrates this concept: each scatter plot represents optimization in the high-dimensional \emph{physical space} of RBF centers and widths, while the central 3D plot depicts the much smaller \emph{hyperparameter space} where Bayesian optimization operates. The latter provides a smooth, interpretable manifold over which kernel distributions evolve toward optimal localization.
	
	\begin{figure}[h]
		\centering
		\includegraphics[width=0.98\textwidth]{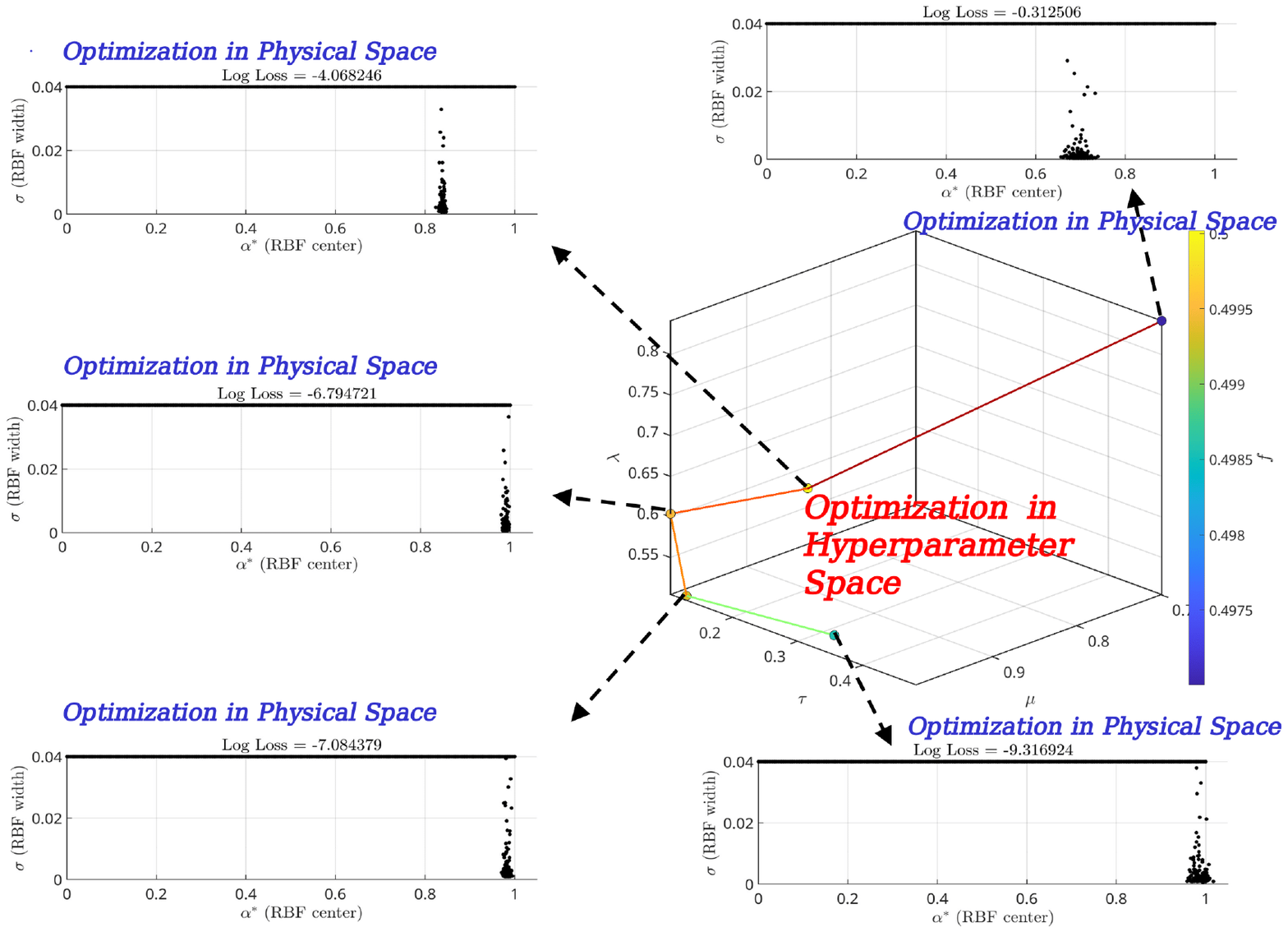}
		\caption{
			Visualization of dimensionality reduction in KAPI-ELM. 
			Multiple physical configurations of RBF centers and widths (outer scatter plots) are governed by a few interpretable distributional hyperparameters $(f,\mu,\tau,\lambda)$ in the low-dimensional hyperparameter space (center). 
			The optimization trajectory in this reduced space drives the adaptive RBF placement observed in physical space.}
		\label{Fig:BL_Interpretation}
	\end{figure}

	\item \textit{Adaptive accuracy with low wall-clock cost.}
	The nested optimization (analytic least squares for $\mathbf{c}$, Bayesian optimization for $\boldsymbol{w}$) achieves high accuracy with modest compute. In the single BL problems, $\log_{10}(J_{\min})<-8$ with wall-clock times of only a few seconds; the twin BL problems also maintain $\log_{10}(J_{\min})<-8$ with wall-clock times under two minutes. For the 2D Poisson case at $\nu=10^{-2}$, Bayesian optimization converged in 119 evaluations ($\approx 61.3$\,s) to $\boldsymbol{w}^*=[0.7723,\,0.4915,\,0.5068,\,0.1554,\,0.8869]$, yielding an $L^2$ error of $4.96\times 10^{-8}$ and residuals below $10^{-4}$—all without mesh refinement or backpropagation.
	
	\item \textit{Automatic localization consistent with physics.}
	Across stiffness levels, the learned $(\mu,\tau)$ (and their symmetric counterparts in the twin BL) moved toward the boundary-layer regions and shrank in scale as $\nu$ decreased, automatically clustering RBF centers where gradients are steep. This behavior mirrors the analytical boundary-layer scaling and is achieved solely through distributional search over $\boldsymbol{w}$.

	\item \textit{Comparison with State-of-the-Art Surrogate Models} 
	\begin{itemize}
		\item \textit{Benchmark Landscape.} A recent comprehensive study by \citet{DEFLORIO2024115396} evaluated four 
		representative surrogate modeling approaches---Physics-Informed Neural Networks (PINNs), Physics-Informed Extreme Learning Machines (PIELM), Deep Theory of 
		Functional Connections (Deep--TFC), and Extreme Theory of Functional Connections (X--TFC)---on the same steady 1D advection--diffusion equation for $\nu\in\{10^{-1},10^{-2},10^{-3},10^{-4}\}$. Their results establish a clear hierarchy of performance. For the mild case $\nu=10^{-1}$, all methods remain accurate, with mean absolute errors ranging from $\mathcal{O}(10^{-5})$ (PINNs) to  $\mathcal{O}(10^{-16})$ (X--TFC).  As $\nu$ decreases, however, the problem becomes increasingly stiff and most methods deteriorate: PINNs fail at $\nu=10^{-2}$ due to spectral bias and training stiffness; Deep--TFC becomes biased by $\nu=10^{-2}$ and fails by $\nu=10^{-3}$; PIELM remains accurate to $\mathcal{O}(10^{-8})$ at $\nu=10^{-3}$; and only X--TFC remains viable in the most challenging regime $\nu=10^{-4}$, achieving errors near $\mathcal{O}(10^{-3})$. This study therefore identifies X--TFC as the strongest existing method for 
		high-Péclet transport in one dimension.
		\item \textit{Comparison with X--TFC.} Figure~\ref{fig:xtfc_kapi} compares the kernel distributions produced by X--TFC and KAPI--ELM for the most challenging case $\nu=10^{-4}$. X--TFC employs a fixed, non-adaptive distribution of neurons with widths that shrink linearly as the network size increases. KAPI--ELM, in contrast, employs a distributional parameterization in which the RBF widths contract exponentially with respect to $\nu$, enabling \emph{sharper and more physically consistent localization} of basis functions in stiff regions. Quantitatively, X--TFC achieves mean absolute errors of order $10^{-3}$ using $\mathcal{O}(10^{4})$ neurons, whereas KAPI--ELM attains $\mathcal{O}(10^{-6})$ accuracy with only $\mathcal{O}(10^{3})$ adaptively distributed kernels. This performance gap reinforces the importance of kernel adaptivity in singularly perturbed regimes.
		\item \textit{Asymptotic relation between X--TFC and KAPI--ELM.} An interesting structural insight arises in the asymptotic limit where the 
		number of localized sharp gradients tends to infinity. In this limit the adaptive kernel ensemble of KAPI--ELM becomes increasingly dense, and the distributional optimization converges toward a uniform configuration with progressively smaller kernel widths.  This behavior replicates the initialization strategy used in X--TFC. Thus, X--TFC's initialization may be interpreted as a \emph{limiting case} of KAPI--ELM corresponding to infinitely fine spatial localization without adaptive reallocation of kernels. KAPI--ELM therefore generalizes the core idea of X--TFC's initialization by allowing the kernel placement and widths to reorganize dynamically in response to the physics of the PDE.
		
	\end{itemize}
	
	\begin{figure}[h]
		\centering
		\includegraphics[width=0.9\textwidth]{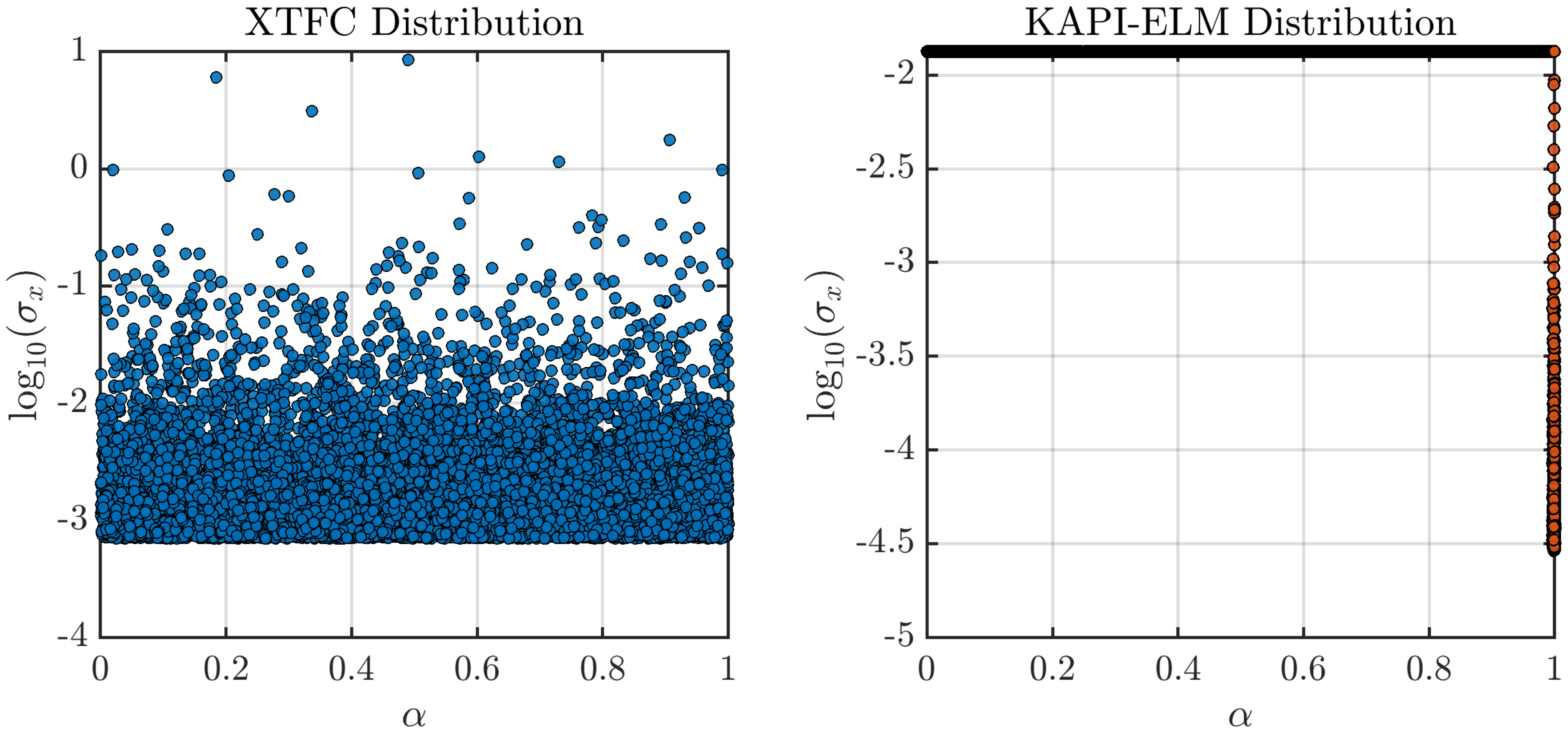}
		\caption{
			Comparison between RBF kernel distributions of X-TFC (deterministic refinement) and KAPI-ELM (distributional adaptivity) for singularly perturbed equations.}
		\label{fig:xtfc_kapi}
	\end{figure}
\end{enumerate}

\subsection{Inverse Problems}

\subsubsection{Problem Setup: 1D Convection-Diffusion}
We next evaluate KAPI-ELM for the inverse identification of the diffusion coefficient~$\nu$ from sparse, noisy measurements, while enforcing the PDE structure referring to equation \ref{eq:conv_diff}.  
The unknown physical parameter~$\nu$ is treated as a deterministic scalar appended to the distributional hyperparameters~$\boldsymbol{w}$, forming the composite optimization vector
\[
\vartheta = [\,\boldsymbol{w};\,\nu\,].
\]
For each trial $\vartheta$, the inner least-squares problem solves for the RBF output weights~$\mathbf{c}^*(\vartheta)$ that minimize the PDE residual.  
The outer Bayesian optimization minimizes the composite objective
\[
J_{\mathrm{inv}}(\vartheta)
= \|\hat{\mathbf{u}}_{\mathrm{obs}}(\vartheta)-\mathbf{u}_{\mathrm{obs}}\|_2^2
+ \lambda_{\mathrm{pde}}\|H_{\mathrm{val}}(\vartheta)\mathbf{c}^{*}(\vartheta)-\mathbf{r}_{\mathrm{val}}\|_2^2,
\]
which balances measurement fidelity with physics consistency.  
Bounds for both distributional and physical parameters are defined as:
\[
\begin{aligned}
	w_f &\in [0.495,\,0.505], &
	w_\mu &\in [0.90,\,0.99], \\[3pt]
	w_\tau &\in [0.10,\,0.40], &
	w_\lambda &\in [0.50,\,1.00], \\[3pt]
	\nu &\in [10^{-4},\,10^{-2}]. &
\end{aligned}
\]

All inverse experiments employ $N_{\mathrm{unif}}=1500$ collocation points and $\eta=0.1$.  
Synthetic data are generated from the forward convection–diffusion problem at $\nu_{\mathrm{true}}=5\times10^{-3}$ with additive Gaussian noise levels of 0\%, 1\%, and 5\%.

\subsubsection{Results and Discussion}
Table~\ref{Tab:InverseResults} lists the optimized parameters for all noise levels.  
KAPI-ELM consistently recovers the true diffusion coefficient within 3\% relative error, maintaining $\log_{10}(J_{\min})<-1.7$ even at 5\% noise.  
The adaptive distributional parameters $(f,\mu,\tau,\lambda)$ remain within physically interpretable ranges, indicating that the model preserves kernel localization consistent with the underlying PDE physics.

\begin{table}[h!]
	\centering
	\caption{Recovered parameters for inverse identification of $\nu_{\mathrm{true}}=0.005$ under varying noise levels.}
	\label{Tab:InverseResults}
	\renewcommand{\arraystretch}{1.15}
	\setlength{\tabcolsep}{5pt}
	\begin{tabular}{|c|c|c|c|c|c|c|c|}
		\hline
		Noise & $f$ & $\mu$ & $\tau$ & $\lambda$ & $\nu_{\mathrm{rec}}$ & $\log_{10}(J_{\min})$ & Time (s) \\
		\hline
		0.0  & 0.4956 & 0.9872 & 0.3063 & 0.5821 & 0.004997 & $-3.83$ & 27.36 \\
		0.01 & 0.4970 & 0.9723 & 0.2840 & 0.8328 & 0.004989 & $-2.42$ & 27.43 \\
		0.05 & 0.5026 & 0.9524 & 0.3974 & 0.5825 & 0.004877 & $-1.73$ & 26.66 \\
		\hline
	\end{tabular}
\end{table}

\begin{figure}[h]
	\centering
	\includegraphics[width=0.95\textwidth]{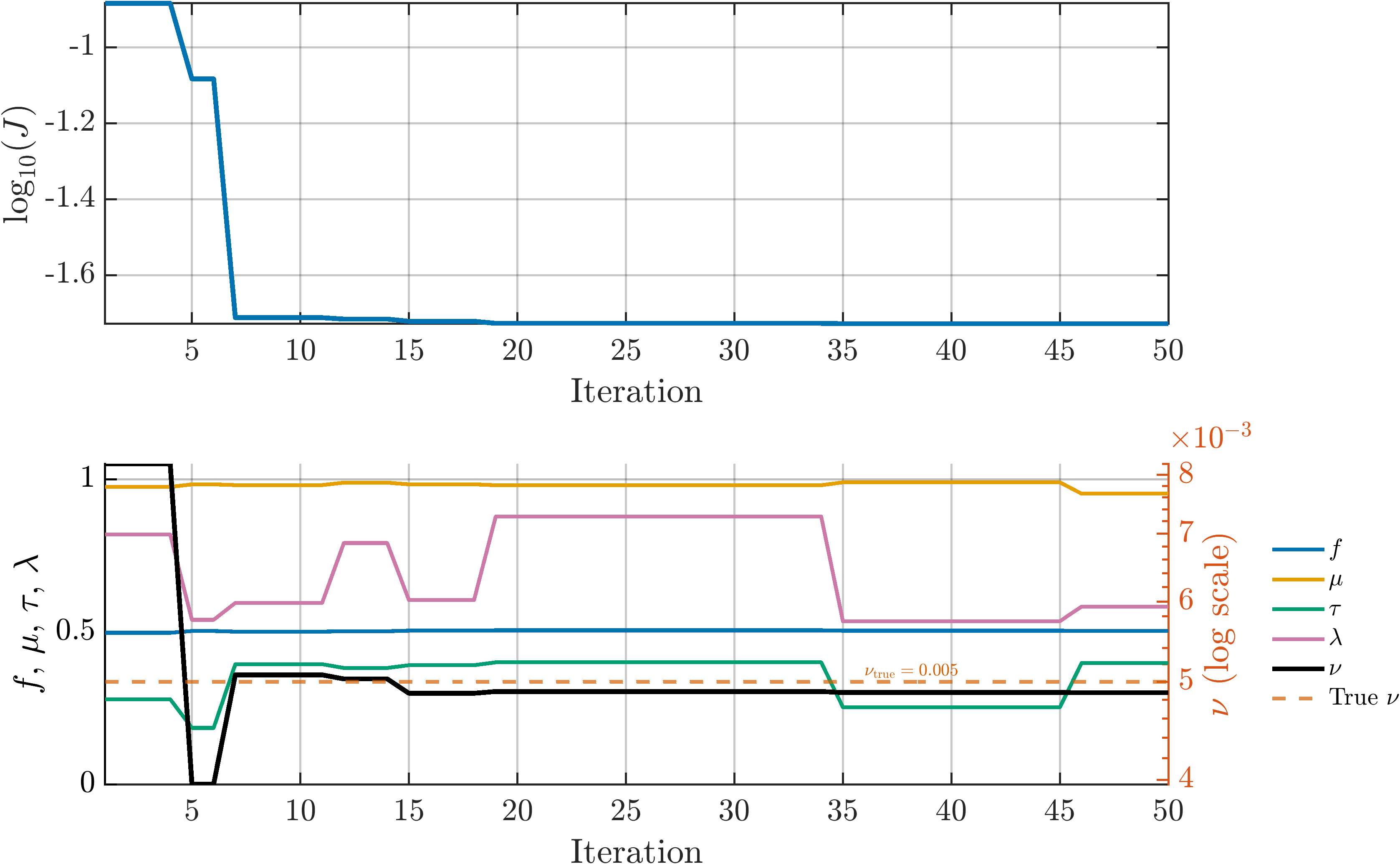}
	\caption{
		Inverse identification of $\nu$ with $5\%$ observation noise.  
		Top: convergence of the best-so-far loss $\log_{10}(J)$.  
		Bottom: evolution of best-so-far parameters $(f,\mu,\tau,\lambda,\nu)$.  
		The recovered $\nu$ (black line) aligns closely with the reference $\nu_{\mathrm{true}}=5\times10^{-3}$ (red dashed line), showing rapid and stable convergence.}
	\label{Fig:KAPI_Training_History}
\end{figure}

Figure~\ref{Fig:KAPI_Training_History} shows the evolution of the best-so-far objective and parameters during Bayesian optimization for the $5\%$ noise case.  
The upper panel displays $\log_{10}(J)$, which rapidly converges within the first 10 iterations.  
The lower panel shows smooth trajectories of the hyperparameters $(f,\mu,\tau,\lambda)$ and the inferred $\nu$ (black curve, right~$y$-axis), which quickly stabilizes near the red dashed reference $\nu_{\mathrm{true}}=5\times10^{-3}$.  
The monotonic loss decay and non-oscillatory parameter evolution indicate a well-conditioned surrogate landscape.

\begin{figure}[h]
	\centering
	\includegraphics[width=0.75\textwidth]{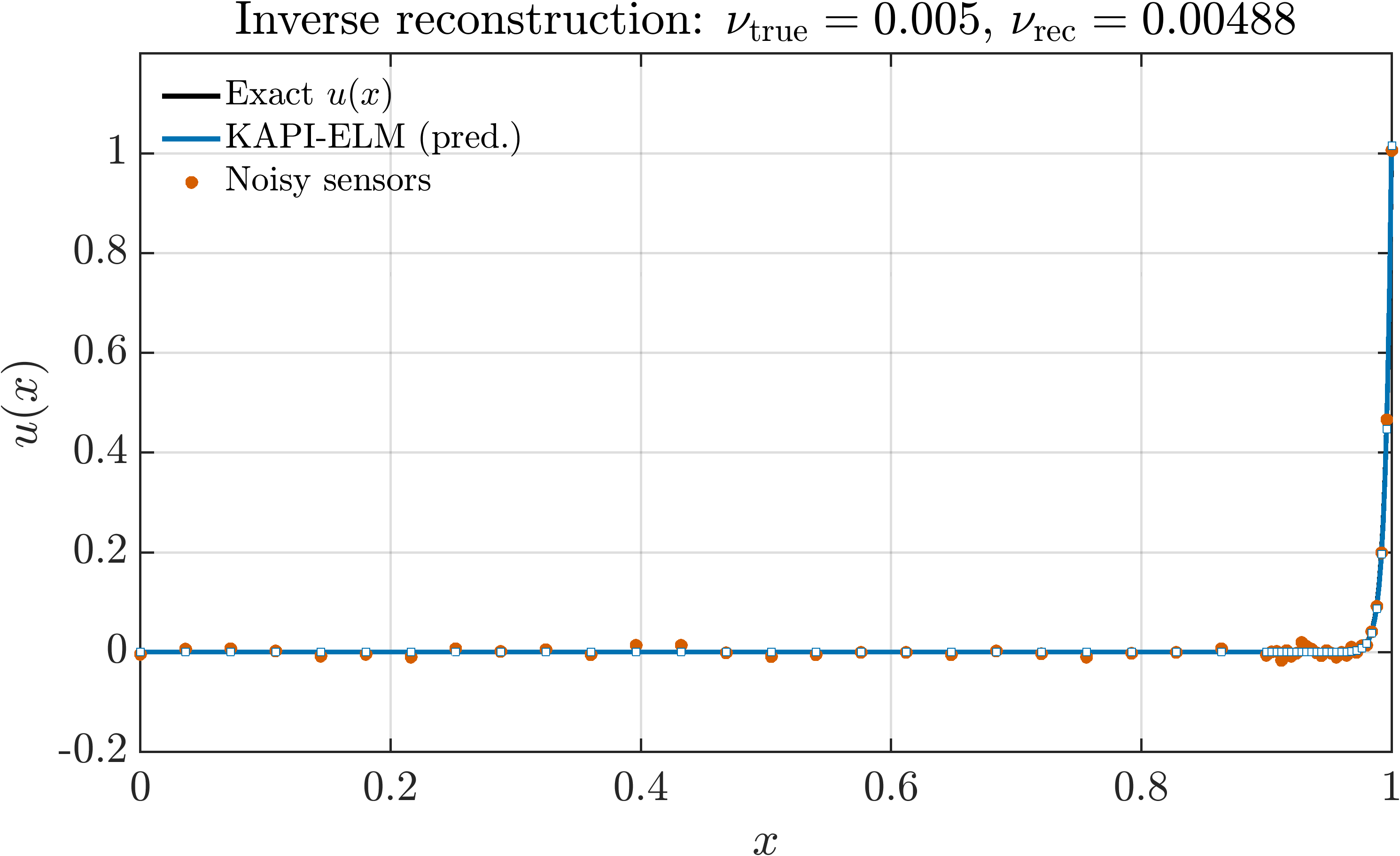}
	\caption{
		Comparison of KAPI-ELM prediction, exact solution, and noisy sensor data ($\nu_{\mathrm{true}}=5\times10^{-3}$, noise = 5\%).  
		KAPI-ELM achieves accurate reconstruction despite sparse, noisy measurements.}
	\label{Fig:SensorVsExact}
\end{figure}

To validate the reconstruction, Figure~\ref{Fig:SensorVsExact} compares the exact field $u(x)$, the noisy sensor data, and the KAPI-ELM prediction.  
Even at 5\% noise, the recovered field matches the ground truth with negligible deviation, confirming accurate physical reconstruction from sparse data.

\subsection*{Remarks}
\begin{enumerate}
	\item \textit{Accurate and stable recovery.}
	KAPI-ELM recovers $\nu$ with $<3\%$ relative error across all noise levels, showing logarithmic degradation of $\log_{10}(J_{\min})$ with increasing noise—consistent with least-squares theory.
	
	\item \textit{Unified optimization of physical and kernel parameters.}
	By jointly optimizing $\nu$ and $\boldsymbol{w}$, the model co-adapts kernel localization and physical parameter inference, improving identifiability without requiring backpropagation.
	
	\item \textit{Efficiency and robustness.}
	Each inverse problem converges within 30\,s on a CPU, demonstrating that the Bayesian outer loop and analytic inner solver provide both speed and numerical stability.
	
	\item \textit{Noise-resilient convergence.}
	The smooth, monotonic loss decay (Fig.~\ref{Fig:KAPI_Training_History}) indicates that the Gaussian-process surrogate regularizes the optimization, mitigating overfitting to noisy data.
	
	\item \textit{Physical consistency.}
	The optimal kernel distributions remain consistent with the forward-calibrated configuration, ensuring that the inverse model captures genuine physical structure rather than data artifacts.
\end{enumerate}

\subsubsection{Problem Setup: 2D Poisson}

We now evaluate the KAPI--ELM framework for inverse estimation of the diffusion
coefficient~$\nu$ in the two-dimensional Poisson equation with a Gaussian source
term.  Synthetic ground--truth fields are generated using a second-order
finite-difference solver on a $100\times 100$ grid with the true parameter
$\nu_{\mathrm{true}}=0.05$.  Observations are sampled at prescribed sensor
locations and corrupted with additive Gaussian noise at levels of $0\%$, $1\%$,
and $5\%$.

As in the one-dimensional convection--diffusion case, the inverse problem is
posed by augmenting the distributional hyperparameters of the adaptive RBF
representation with the unknown physical coefficient~$\nu$.  The composite
optimization vector is
\[
\vartheta = \bigl[\, f,\,\mu_x,\,\mu_y,\,\tau,\,\lambda;\,\nu \,\bigr].
\]

The admissible ranges of the distributional parameters and the physical
coefficient mirror those used in the forward solver:
\[
\begin{aligned}
	f_{\mathrm{adap}} &\in [0.5,\,1.0], &
	\mu_x,\mu_y &\in [0.2,\,0.8], \\[3pt]
	\tau &\in [0.1,\,0.5], &
	\lambda &\in [0.5,\,0.9], \\[3pt]
	\nu &\in [5\times10^{-3},\,5\times10^{-1}]. &
\end{aligned}
\]
All experiments employ a uniform collocation grid with $N_{\mathrm{unif}}=40^2$
PDE residual points and a maximum RBF width parameter $\sigma_{\max}=0.2$.

\subsubsection{Results and Discussion}

Table~\ref{Tab:InverseResults2D} reports the optimal parameters recovered for all
noise levels.  For noise-free data, KAPI--ELM identifies the physical parameter
with a relative error below $0.3\%$, yielding $\nu_{\mathrm{rec}}=0.05016$.
Under $1\%$ and $5\%$ noise, the recovered values remain close to the ground
truth, with $\nu_{\mathrm{rec}}=0.05134$ and $\nu_{\mathrm{rec}}=0.04691$,
respectively. 

\begin{figure}[h!]
	\centering
	\begin{tabular}{c}
		\includegraphics[width=0.92\textwidth]{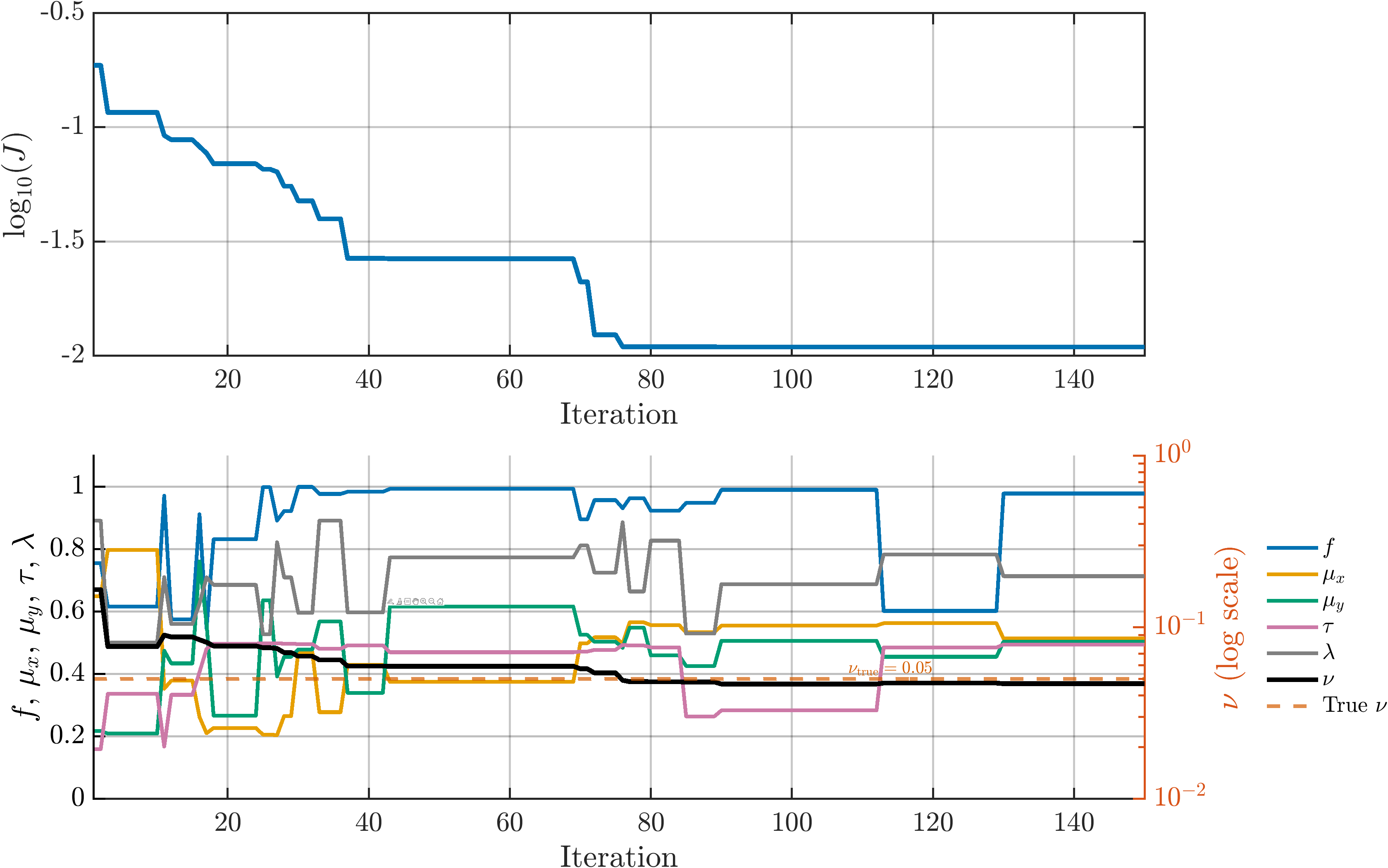} \\[8pt]
		\includegraphics[width=0.92\textwidth]{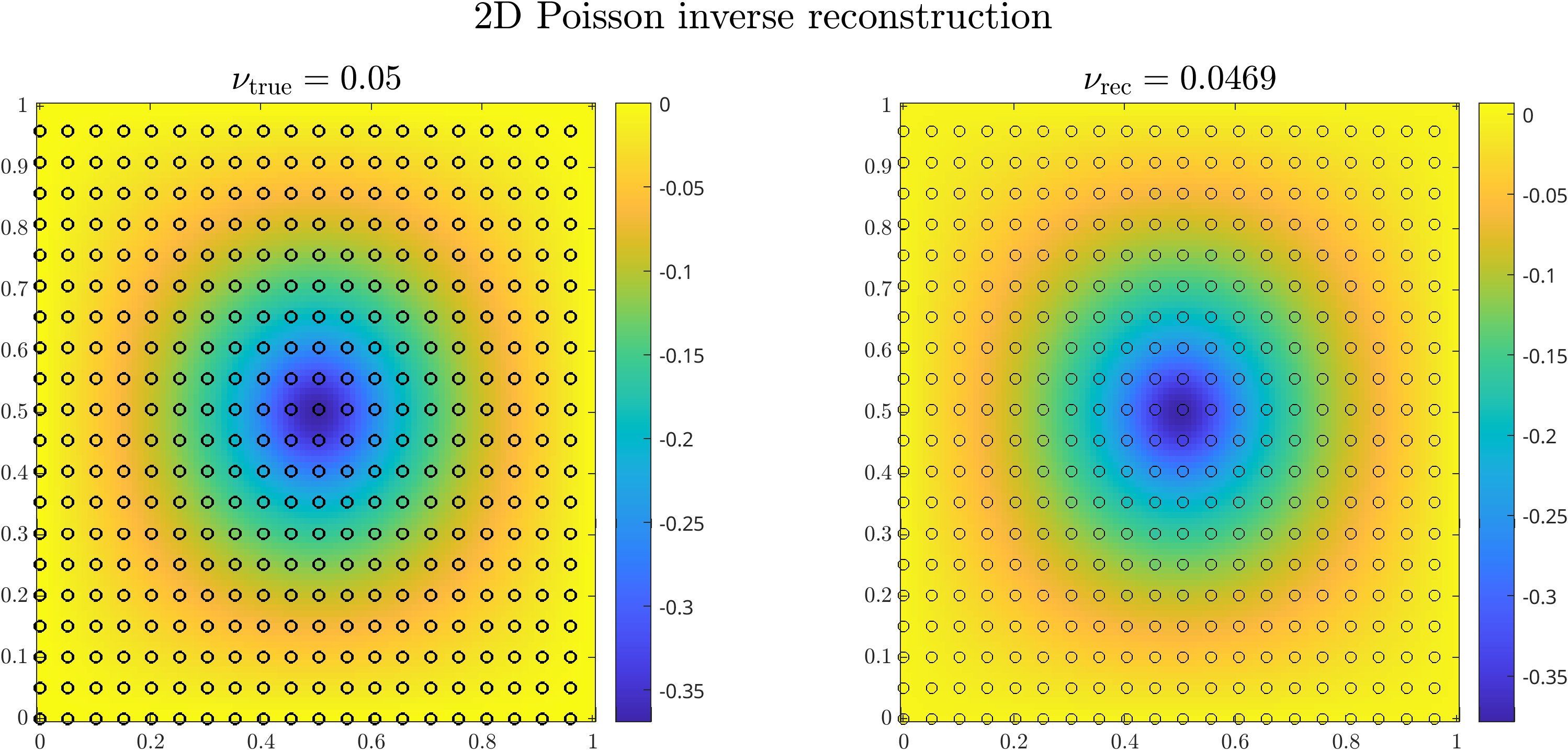}
	\end{tabular}
	\caption{
		Inverse identification for the 2D Poisson problem with $5\%$ observation noise.  
		\textbf{Top:} Best-so-far Bayesian optimization history, showing monotonic decay of the inverse objective $\log_{10}(J)$ and stable convergence of the distributional parameters $(f_{\mathrm{adap}},\mu_x,\mu_y,\tau,\lambda)$ along with the inferred diffusion coefficient~$\nu$ (black curve).  
		\textbf{Bottom:} Comparison of the exact field generated with $\nu_{\mathrm{true}}=0.05$ and the KAPI--ELM reconstruction using the optimized parameters, which yields $\nu_{\mathrm{rec}}=0.0469$.  
		The reconstructed field closely matches the ground truth, demonstrating accurate and noise-robust inverse recovery.
	}
	\label{Fig:2x1PoissonInverse}
\end{figure}

The evolution of the Bayesian optimization process is shown in
Fig.~\ref{Fig:2x1PoissonInverse} (top panel). Figure~\ref{Fig:2x1PoissonInverse} (bottom panel) compares the reconstructed
KAPI--ELM field with the finite-difference ground truth for the $5\%$ noise
scenario.  
The solution faithfully captures the localized potential well generated by the
Gaussian source, preserving both the amplitude and spatial structure of the
field.  
Despite measurement noise, the reconstructed solution remains visually
indistinguishable from the exact field, demonstrating that KAPI--ELM achieves
robust and accurate inverse inference in two-dimensional PDE settings while
maintaining strong adherence to the underlying physics.

\begin{table}[h!]
	\centering
	\caption{Recovered parameters for inverse identification of
		$\nu_{\mathrm{true}}=0.05$ in the 2D Poisson problem.}
	\label{Tab:InverseResults2D}
	\renewcommand{\arraystretch}{1.15}
	\setlength{\tabcolsep}{5pt}
	\begin{tabular}{|c|c|c|c|c|c|c|c|}
		\hline
		Noise & $f_{\mathrm{adap}}$ & $\mu_x$ & $\mu_y$ &
		$\tau$ & $\lambda$ & $\nu_{\mathrm{rec}}$ & $\log_{10}(J_{\min})$ \\
		\hline
		0.0  & 0.9364 & 0.5110 & 0.5065 & 0.4722 & 0.7564 & 0.05016 & $-3.05$ \\
		0.01 & 0.6254 & 0.4600 & 0.5022 & 0.4643 & 0.5329 & 0.05134 & $-2.35$ \\
		0.05 & 0.9785 & 0.5136 & 0.5035 & 0.4948 & 0.7137 & 0.04691 & $-1.96$ \\
		\hline
	\end{tabular}
\end{table}

\subsection{Lid-Driven Cavity Flow: Kernel Adaptivity with Curriculum Learning}
\label{sec:lid-driven}

The distributional optimization strategy in KAPI-ELM is not restricted to resolving sharp gradients. When paired with curriculum learning~\cite{DWIVEDI2025130924}, it can also be extended to handle nonlinear PDEs, as discussed in this section.

\subsubsection{Model Problem } 
We consider the two-dimensional, steady, incompressible Navier-Stokes equations on the unit square
\[
\Omega = [0,1]^2,\qquad \partial\Omega = \Gamma_L \cup \Gamma_R \cup \Gamma_B \cup \Gamma_T,
\]
governed by
\begin{equation}
	\label{eq:ns_steady}
	\nabla \cdot \mathbf{V} = 0,
	\qquad
	(\mathbf{V}\!\cdot\nabla)\mathbf{V}
	= -\,\nabla p + \frac{1}{Re}\,\nabla^{2}\mathbf{V},
\end{equation}

where $\mathbf{V}=(u,v)^T$ is the velocity, $p$ is the pressure, and $Re$ denotes the Reynolds number.

We impose a lid-driven cavity configuration: no-slip on the side and bottom walls and a unit tangential lid speed at the top,
\begin{equation}
	\boldsymbol{u} = (0,0)\ \text{on}\ \Gamma_L\cup\Gamma_R\cup\Gamma_B,
	\qquad
	\boldsymbol{u} = (1,0)\ \text{on}\ \Gamma_T,
	\label{eq:stokes_bcs}
\end{equation}
with a pressure reference at a corner to fix the additive constant,
\begin{equation}
	p(0,0) = 0.
	\label{eq:stokes_p_ref}
\end{equation}

The underlying idea is to replace the full nonlinear Navier-Stokes system with a \emph{sequence of linearized problems} obtained by gradually increasing the Reynolds number. At each stage, the nonlinear advection term is frozen using the velocity field from the previous stage, yielding the linearized approximation
\[
(\mathbf{V}\!\cdot\nabla)\mathbf{V}\big|_{Re}
\;\approx\;
(\mathbf{V}_{Re-\delta}\!\cdot\nabla)\mathbf{V}_{Re},
\]
with $\mathbf{V}_{Re-\delta}$ denoting the flow solution from the preceding stage. This procedure converts the nonlinear Navier-Stokes equations into a sequence of linear least-squares problems, each fully compatible with the analytic PI-ELM structure.

\subsubsection{PI-ELM Ansatz}
Unlike PINNs that rely on multilayer networks trained via backpropagation, PI-ELMs employ a single hidden-layer feedforward architecture with fixed input parameters, resulting in a linear dependence on the output weights. 
For the Stokes system, we approximate each field component $q\in\{u,v,p\}$ as a global weighted sum of Gaussian radial basis functions,
\[
\hat{q}(x,y)
= \sum_{k=1}^{N^*} c_{q,k}\,\phi\!\big(z_k(x,y)\big),
\qquad 
\phi(z) = e^{-z^2},
\]
where
\[
z_k(x,y)
= \sqrt{(m_k x + \alpha_k)^2 + (n_k y + \beta_k)^2}.
\]
The parameters $(m_k,n_k,\alpha_k,\beta_k)$ define the scale and translation of each basis function, shared across $(u,v,p)$ to preserve pressure–velocity coupling~\cite{DWIVEDI2025130924}. This shared RBF configuration ensures that all three fields are represented within a common adaptive kernel basis, enabling consistent enforcement of both momentum and continuity equations. Our target Reynolds numbers are 0.01, 10, and 100.

\subsubsection{Distributional Hyperparameters}
\label{sec:ns_dist_hyp}
The adaptive kernel layout in the Navier--Stokes experiments is governed by a four–dimensional hyperparameter vector
\[
\boldsymbol{w} = (w_1,\,w_2,\,w_3,\,w_4),
\qquad
w_1\in[0.5,1.0],\;
w_2,w_3,w_4\in[0.1,0.5].
\]
These parameters jointly determine the density of collocation points, the
number of RBF centers, and the spatial variation of kernel widths.

For a given $w_1$, the Chebyshev--Lobatto resolution is
\[
N_x(w_1) = \lfloor 30\, w_1 \rfloor,
\] where the notation $\lfloor \cdot \rfloor$ represents the floor function, which returns the greatest integer less than or equal to the argument.

This results in a total of
\begin{equation}
	N_c(w_1)
	=
	\underbrace{\bigl(N_x(w_1)-2\bigr)^2}_{\text{interior}}
	\;+\;
	\underbrace{4\bigl(N_x(w_1)-2\bigr)}_{\text{boundary}}
	\label{eq:Nc_expression}
\end{equation}

collocation points.  
Given this grid resolution, the number of RBF centers is
\begin{equation}
	N_{\mathrm{rbf}}(w_1,w_2)
	=
	\left\lfloor 
	\sqrt{\, w_2 \bigl(N_x(w_1)-2\bigr)^2 }
	\right\rfloor^{2}.
	\label{eq:Nrbf_expression}
\end{equation}
Across the search ranges, these expressions yield approximately
$N_c\approx 221$--$896$ collocation points and
$N_{\mathrm{rbf}}\approx 16$--$400$ RBF centers.

\paragraph{Physical interpretation}

\begin{enumerate}
	\item The parameter $w_1$ controls the density of the Chebyshev--Lobatto grid and
	thus the enforcement strength of the PDE and boundary constraints; larger
	values correspond to finer spatial resolution and a more accurate capture of
	boundary layers.  
	\item The parameter $w_2$ regulates the global density of RBF centers and therefore
	the representational capacity of the basis; smaller $w_2$ produces a denser
	kernel ensemble with narrower supports, while larger values favor coarser,
	more global kernels.
	\item The remaining parameters $(w_3,w_4)$ control how the Gaussian widths vary with
	the distance of each RBF center to the cavity boundary.  Let $d_i$ denote the
	minimum distance of center $i$ to the walls.  The width is modeled as
	\[
	\sigma_i = w_4 + w_3\, d_i,
	\qquad
	w_3,\,w_4 \in [0.1,\,0.5].
	\]
	Here, $w_3$ governs the \emph{rate} at which the kernel width increases from
	the walls toward the interior.  Larger $w_3$ values create a sharp contrast
	between narrow kernels near the walls—well suited for resolving thin shear
	layers and corner vortices—and broader kernels in the interior.  
	The parameter $w_4$ sets the \emph{baseline} or minimum RBF width attained at
	the boundary.  Smaller $w_4$ allows highly localized kernels, improving the
	resolution of steep gradients, whereas larger values produce smoother, more
	globally supported kernels that may enhance conditioning.  
	Together, $(w_3,w_4)$ provide a simple yet effective wall-aware scaling law
	that allocates resolution where the Navier--Stokes operator demands it most,
	particularly as the Reynolds number increases.
\end{enumerate}

\subsubsection{Residual Expressions}
\begin{itemize}
	\item \textit{PDE residual}: For each collocation point within $\Omega$, residuals are calculated from continuity, $x$ and $y$- momentum equations and set them equal to zero as follows:
	\begin{equation}
		\boldsymbol{M}\overrightarrow{c}=\overrightarrow{0}
	\end{equation}
	where $\boldsymbol{M}$ is a block matrix and $\overrightarrow{c}$ is outer layer weights matrix. Specifically,
	\begin{equation}
		\left[\begin{array}{ccc}
			M_{u1} & M_{v1} & M_{p1}\\
			M_{u2} & M_{v2} & M_{p2}\\
			M_{u3} & M_{v3} & M_{p3}
		\end{array}\right]\left(\begin{array}{c}
			\overrightarrow{c_{u}}\\
			\overrightarrow{c_{v}}\\
			\overrightarrow{c_{p}}
		\end{array}\right)=\left(\begin{array}{c}
			0\\
			0\\
			0
		\end{array}\right)
	\end{equation}
	Here the shape of individual blocks is $1 \times N^{*}$. The first row corresponds to the continuity equation. The second and third rows correspond to the $x$ and $y$- momentum equations, respectively. For $k=1,2,..,N^{*}$,
	\begin{equation}
		M_{u1}(1,k)=-2e^{\xi_{k}}m_{k}(m_{k}x+\alpha_{k})
	\end{equation}
	\begin{equation}
		M_{v1}(1,k)=-2e^{\xi_{k}}n_{k}(n_{k}y+\beta_{k})
	\end{equation}
	\begin{equation}
		M_{p1}(1,k)=0
	\end{equation}
	\begin{multline}
		M_{u2}(1,k)=\hat{u}_{ref}M_{u1}(1,k)+\hat{v}_{ref}M_{v1}(1,k)\\
		+\frac{2e^{\xi_{k}}}{Re}[m_{k}^{2}\{1-2(m_{k}x+\alpha_{k})^{2}\}+n_{k}^{2}\{1-2(n_{k}y+\beta_{k})^{2}\}]
	\end{multline}
	\begin{equation}
		M_{v2}(1,k)=0
	\end{equation}
	\begin{equation}
		M_{p2}(1,k)=M_{u1}(1,k)
	\end{equation}
	\begin{equation}
		M_{u3}(1,k)=0
	\end{equation}
	\begin{equation}
		M_{v3}(1,k)=M_{u2}(1,k)
	\end{equation}
	\begin{equation}
		M_{p3}(1,k)=M_{v1}(1,k)
	\end{equation}
	The terms $\hat{u}_{ref}$ and $\hat{v}_{ref}$ represent the reference velocities used in the quasi-linear approximation of the nonlinear advection terms, as previously discussed.
	
	\item \textit{Boundary condition residual:} For each boundary point in $\partial\Omega$, the residual depends on the boundary type (Dirichlet, Neumann, or mixed). For a no-slip condition,
	\[
	\begin{bmatrix}
		B_{u1} & B_{v1} & B_{p1}\\
		B_{u2} & B_{v2} & B_{p2}
	\end{bmatrix}
	\begin{bmatrix}
		\vec{c_u}\\ \vec{c_v}\\ \vec{c_p}
	\end{bmatrix}
	=
	\begin{bmatrix}0\\0\end{bmatrix},
	\]
	where for $k=1,\dots,N^*$:
	$B_{u1}(1,k)=e^{\xi_k}$, $B_{v1}(1,k)=B_{p1}(1,k)=0$, $B_{u2}(1,k)=0$, $M_{v2}(1,k)=B_{u1}(1,k)$, and $M_{p2}(1,k)=0$. A horizontal velocity inlet is obtained by replacing the RHS with $\begin{bmatrix}U(y)\\0\end{bmatrix}$.
	
\end{itemize}

\subsubsection{Optimization}
Starting from $Re=0$, the formulation simplifies to the standard steady Stokes problem. Subsequently, each curriculum step (here using $\delta=0.1$) re-optimizes the RBF kernel distribution using Bayesian optimization, computes the corresponding flow field through a single closed-form least-squares solve, and then updates the Reynolds number via $Re \leftarrow Re+\delta$. The resulting velocity field is then used as the spatially varying coefficient in the linearized advection term for the next stage. 

\subsubsection{Results}

We assess the accuracy of the curriculum-driven KAPI-ELM solution across a range of Reynolds numbers by comparing (i) centerline velocity profiles with the high-resolution finite-volume data of Marchi et al.~\cite{marchi2009lid} and (ii) pointwise errors in the velocity magnitude relative to a finely resolved FEM reference solution. 

\paragraph{Approximate Meta-Learning Across Reynolds Numbers}
A key advantage of the curriculum-based KAPI--ELM framework is that the distributional hyperparameters optimized during the continuation in $Re$ can be
stored and reused for \emph{any} intermediate Reynolds number $Re' \le 100$.  The initial curriculum sweep up to $Re=100$ requires the full Bayesian-optimization effort and takes approximately $26$ minutes.  However, once these optimal hyperparameters have been identified, subsequent evaluations for any $Re' \le 100$ bypass the optimization entirely and require only the analytic least-squares solve, completing in roughly $1.5$ minutes.

This stands in sharp contrast to PINNs: a PINN trained at $Re=100$ can be used only for inference at that specific parameter value, whereas the KAPI--ELM
hyperparameters act as a reusable kernel-distribution model that generalizes across the entire Reynolds-number range.  The resulting behavior is analogous
to operator-learning frameworks that amortize training across parameter families, but achieved here within a lightweight, gradient-free physics-informed solver.

\begin{figure}[h!]
	\centering
	\includegraphics[width=0.95\linewidth]{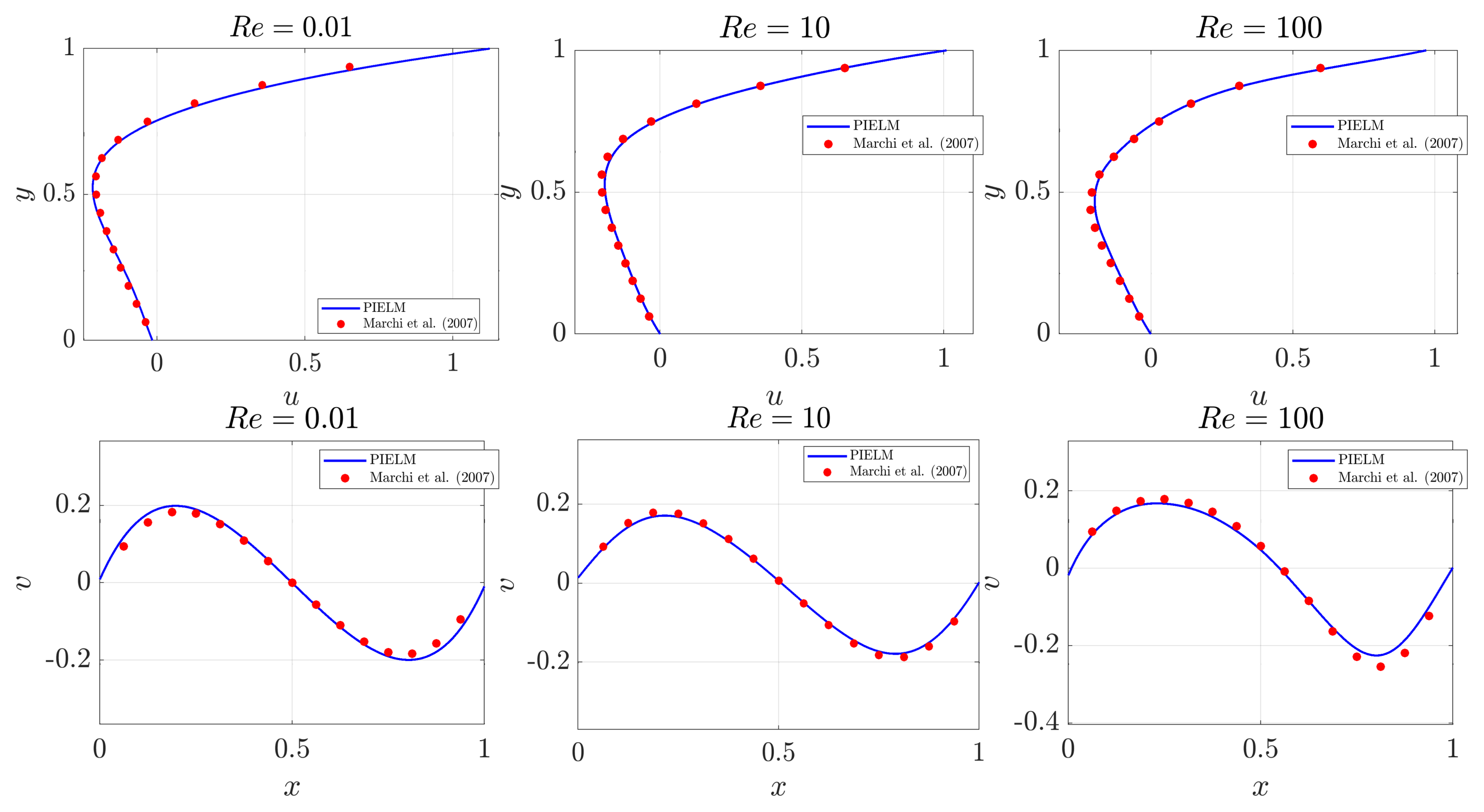}
	\caption{Centerline velocity comparison. 
		Horizontal velocity $u$ along the vertical centerline (top row) and vertical velocity $v$ along the horizontal centerline (bottom row) for $Re=0.01$, $10$, and $100$.  
		KAPI--ELM predictions (blue) exhibit excellent agreement with the high-resolution finite-volume data of Marchi et al.~(2007) (red).}
	\label{fig:centerline_lid}
\end{figure}

\paragraph{Centerline velocity comparison}
Figure~\ref{fig:centerline_lid} reports the horizontal velocity $u$ along the vertical centerline and the vertical velocity $v$ along the horizontal centerline for $Re\in\{0.01,10,100\}$. Across all Reynolds numbers, the KAPI--ELM predictions (blue curves) match the benchmark data of Marchi et al.~\cite{marchi2009lid} (red symbols) with high fidelity. For creeping flow ($Re=0.01$), the agreement is nearly indistinguishable, reflecting the smooth nature of the solution.  
As $Re$ increases, shear layers form near the moving lid and secondary vortical structures intensify; nevertheless, the centerline profiles remain in close agreement with the benchmark data, demonstrating that the curriculum-driven continuation in Reynolds number successfully stabilizes the solution of the nonlinear operator.

\begin{figure}[h!]
	\centering
	\begin{tabular}{ccc}
		\includegraphics[width=0.32\linewidth]{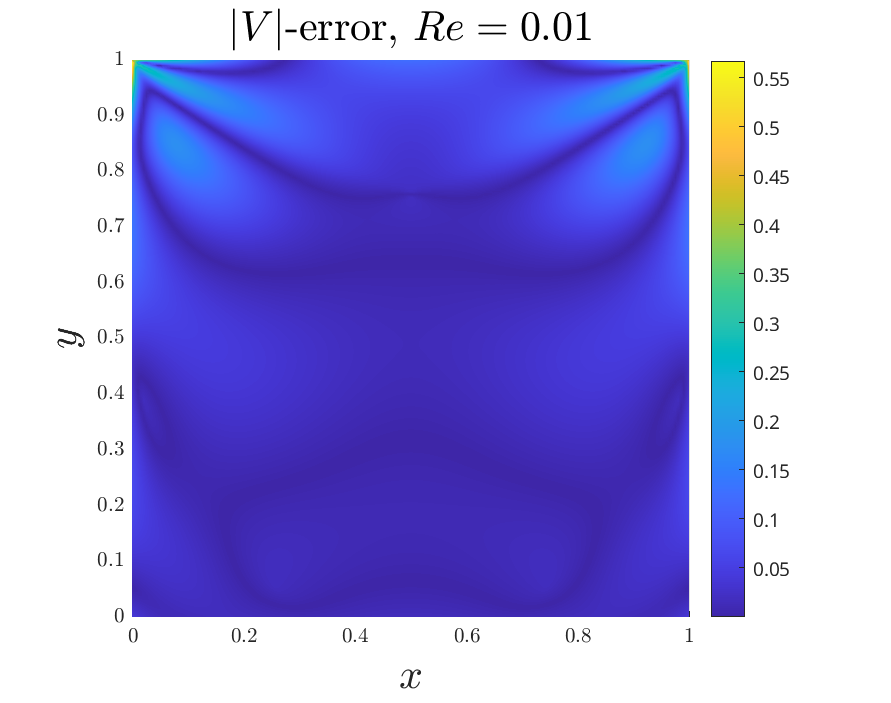} &
		\includegraphics[width=0.32\linewidth]{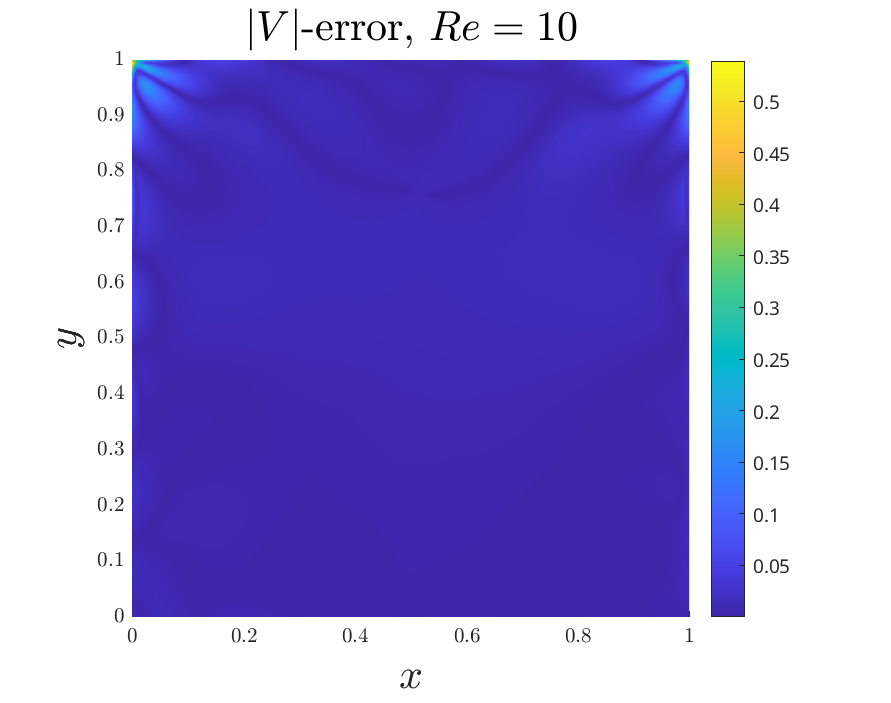} &
		\includegraphics[width=0.32\linewidth]{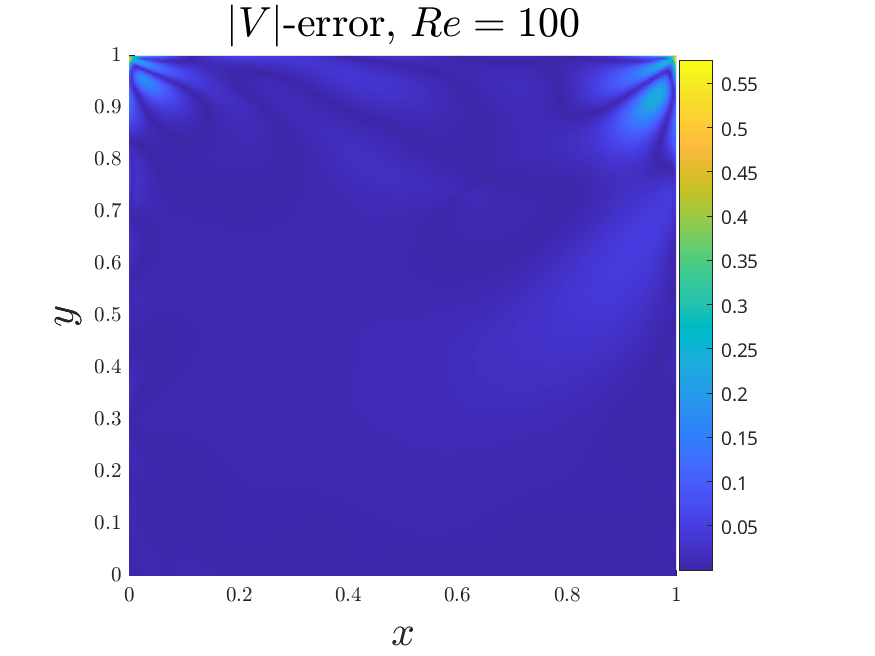}
	\end{tabular}
	\caption{Velocity-magnitude error fields $e(x,y)$ for $Re=0.01$, $10$, and $100$, computed relative to a high-resolution FEM reference.  
		Errors remain uniformly small across the domain, with mild increases near the lid corners where the flow features sharp directional changes.}
	\label{fig:lid_error}
\end{figure}

\paragraph{Pointwise velocity-magnitude errors}
To further quantify accuracy over the full domain, we compute the pointwise error in velocity magnitude,
\[
e(x,y) = \bigl|\,\mathbf{V}_{\mathrm{KAPI-ELM}}(x,y) - \mathbf{V}_{\mathrm{FEM}}(x,y)\,\bigr|,
\]
where $\mathbf{V}_{\mathrm{FEM}}$ denotes a high-resolution finite-element solution used as a surrogate ground truth.  
The error fields for $Re=0.01$, $10$, and $100$ are shown in Fig.~\ref{fig:lid_error}.  
In all cases, the errors remain small throughout the domain, with the largest discrepancies localized near the lid corners where gradients are strongest.  
Importantly, the error levels do not deteriorate significantly as $Re$ increases, indicating that the adaptive kernel distribution continues to track the evolving shear layers effectively.

\begin{figure}[h]
	\centering
	\includegraphics[width=0.95\linewidth]{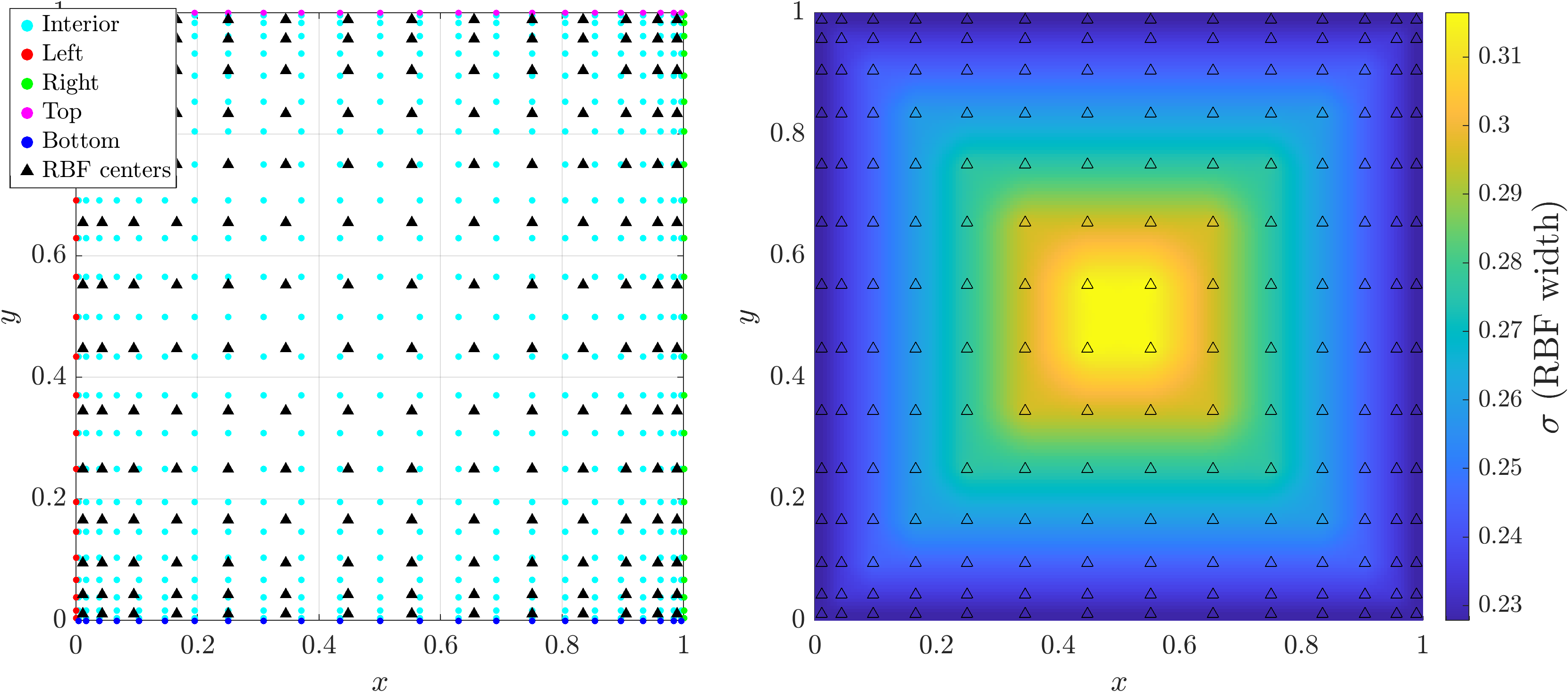}
	\caption{Adaptive collocation points and RBF width field for $Re=10$.  
		Kernel supports adjust to the milder shear-layer structure at this Reynolds level.}
	\label{fig:rbf_Re10}
\end{figure}

\begin{figure}[h]
	\centering
	\includegraphics[width=0.95\linewidth]{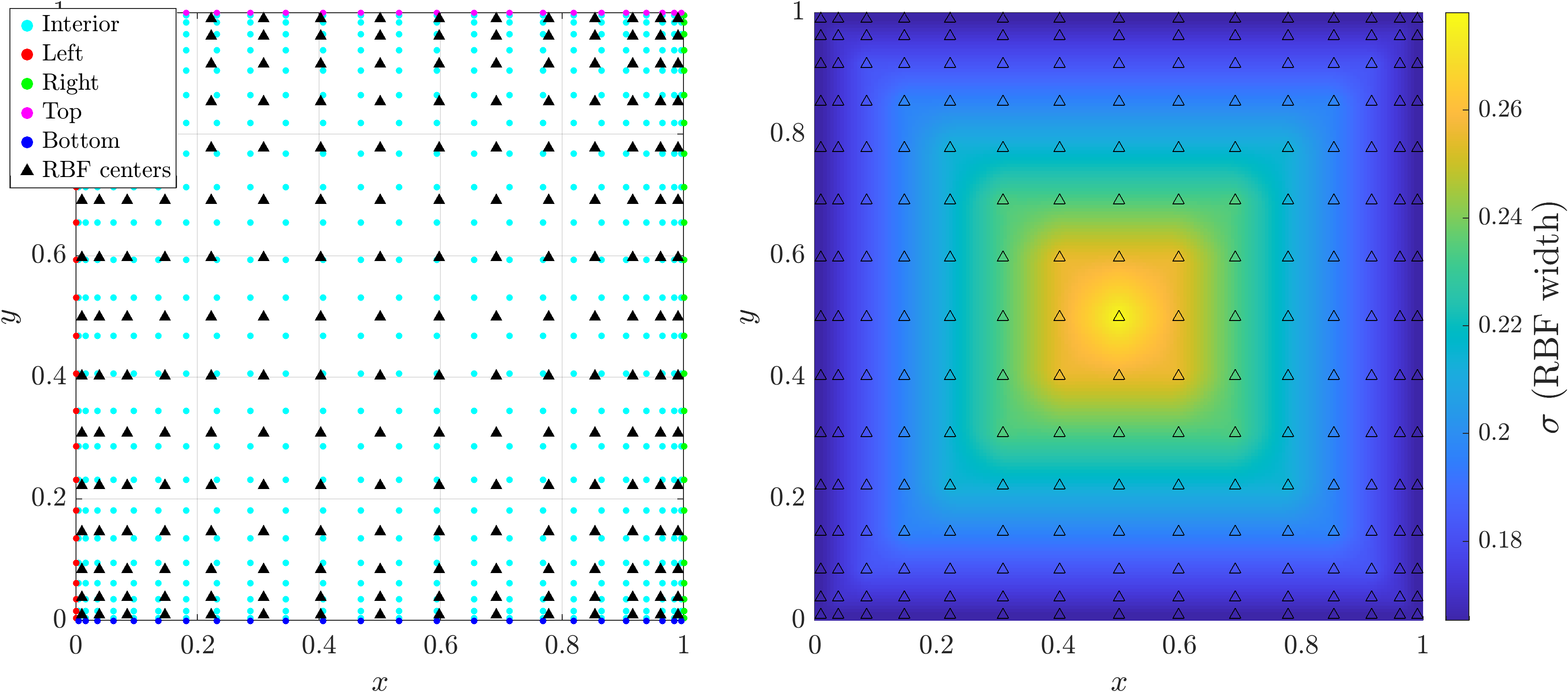}
	\caption{Adaptive collocation points and RBF width field for $Re=100$.  
		Compared with $Re=10$, the kernel distribution tightens near the lid and corners to capture stronger shear gradients.}
	\label{fig:rbf_Re100}
\end{figure}

\begin{figure}[h!]
	\centering
	\includegraphics[width=\linewidth]{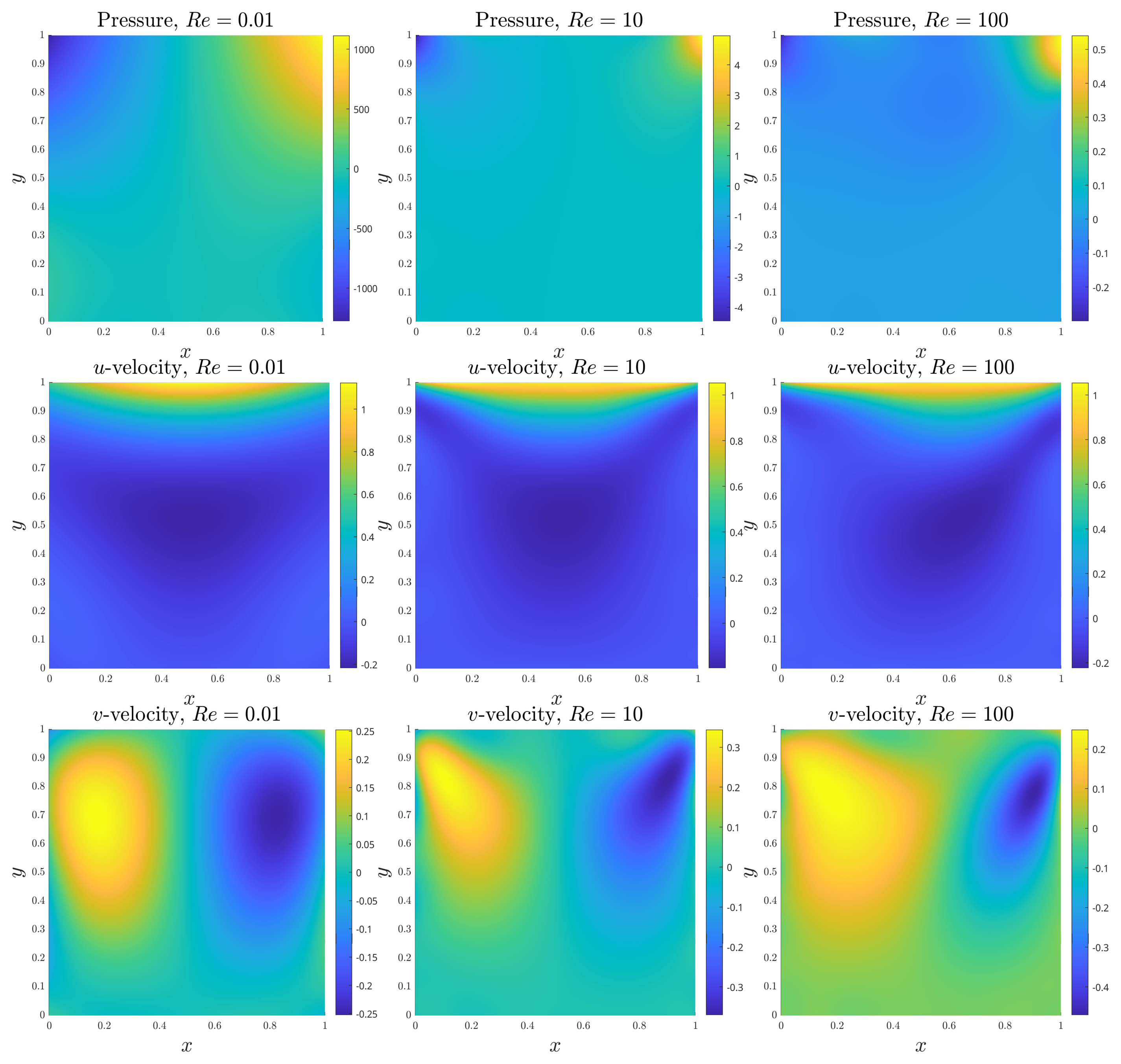}
	\caption{KAPI-ELM predicted pressure and velocity fields for the lid-driven cavity. Shown are the pressure $p$ (top row), horizontal velocity $u$ (middle row), and vertical velocity $v$ (bottom row) for $Re = 0.01$, $10$, and $100$.}
	\label{fig:lid_uvp}
\end{figure}

\paragraph{Adaptive kernel layout}
Because the distributional parameters are re-optimized at every curriculum stage, the RBF centers and their widths evolve with increasing Reynolds number, yielding systematically different kernel configurations at moderate and high $Re$; two representative layouts are shown in Fig.~\ref{fig:rbf_Re10} and Fig.~\ref{fig:rbf_Re100}. KAPI-ELM predicted pressure and velocity fields are shown in Fig.~\ref{fig:lid_uvp}. As the Reynolds number increases, the solution develops stronger shear layers near the moving lid which are accurately captured by the curriculum-driven KAPI-ELM. The smooth transition of flow features across increasing $Re$ highlights both the stability of the curriculum learning strategy and the effectiveness of the adaptive kernel distribution.

Overall, these results demonstrate that the combination of kernel adaptivity and curriculum learning enables KAPI-ELM to robustly and accurately approximate the Navier-Stokes operator across a wide range of Reynolds numbers without resorting to iterative backpropagation or deep-network architectures.

\FloatBarrier
\section{Limitations and Future Scope}
\label{sec:limitations}
Although the proposed KAPI-ELM framework demonstrates strong performance, several limitations remain.  
First, the Bayesian optimization of distributional hyperparameters can become computationally demanding as the number of adaptive components increases, since each evaluation requires solving an embedded least-squares system. More scalable strategies—such as surrogate-assisted search, multi-fidelity Bayesian optimization, or gradient-informed kernel tuning—could alleviate this cost.

Second, the present study focuses exclusively on steady problems. Transient, convection-dominated, or transport-driven systems—where sharp features evolve over time—remain unaddressed. Extending KAPI-ELM to these regimes will require temporal adaptivity, dynamic kernel transport, and potentially joint optimization of spatial and temporal kernel distributions.

Third, although the hyperparameter bounds are physically interpretable, they still rely on informed user choices and limited trial-and-error. Automating this process through hierarchical Bayesian priors or meta-learning could further improve robustness and reproducibility.

\begin{figure}[h]
	\centering
	\includegraphics[width=0.95\linewidth]{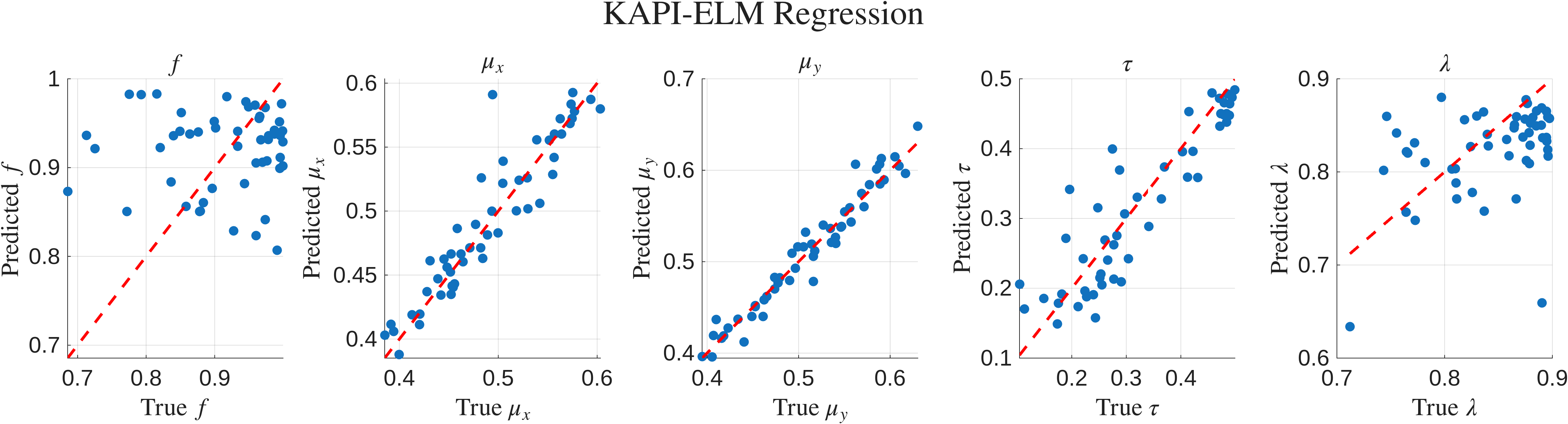}
	\caption{True vs.\ predicted values of the kernel distribution parameters 
		$(f,\mu_x, \mu_y, \tau,\lambda)$ obtained from the regression network. 
		The model is trained to map Gaussian source descriptors $(x_0, y_0, \nu)$ 
		to the optimized kernel parameters used in KAPI--ELM. 
		Each subplot compares ground-truth values from the dataset with neural network 
		predictions on the held-out test set. 
		The strong alignment with the unity line indicates accurate learning and vice versa.}
	
	\label{fig:limit}
\end{figure}

Finally, the current parameterization of the RBF distribution is only one possible design choice and should not be viewed as optimal. For the 2D Poisson problem, for instance, we employed five hyperparameters \([f, \mu_x, \mu_y, \tau, \lambda]\), where \(\{\mu_x, \mu_y, \tau\}\) are geometric length-scale parameters. When these are swept across the PDE parameters like center of source term and \(\nu\), they exhibit strong and intuitive correlations, whereas the dimensionless parameters \(f\) and \(\lambda\) correlate more weakly. This indicates that although the chosen parameterization is physically meaningful, it may not be the most expressive (See Fig. \ref{fig:limit}). Nevertheless, the framework delivers good results even under these sub-optimal choices. We anticipate that future work will explore more principled kernel parameterizations---potentially informed by physics, symmetries, or data-driven structure---which can further enhance the efficiency and predictive capability of the KAPI-ELM methodology.

\section{Conclusion}
\label{Sec:Conclusion}

This work introduced the Kernel-Adaptive Physics-Informed Extreme Learning Machine (KAPI-ELM), a new class of physics-informed solvers that optimizes the \emph{distribution} of radial basis function (RBF) kernels rather than individual network weights. By shifting the search from a high-dimensional parameter space, as in PINNs, to a low-dimensional and physically interpretable hyperparameter space, KAPI-ELM provides a lightweight yet expressive alternative for solving forward and inverse partial differential equations.

The proposed method retains the defining strengths of the PI-ELM framework—analytic least-squares training, fast convergence, and architectural simplicity—while addressing its primary limitation: the fixed nature of kernel centers and widths. Through Bayesian optimization of kernel density and support, KAPI-ELM adaptively concentrates resolution where the PDE solution develops high PDE residuals, without relying on backpropagation and without altering the linear training structure.

Extensive numerical experiments demonstrate the versatility of the approach. In the forward setting, KAPI-ELM accurately resolves sharp interior and boundary layers in one-dimensional convection-diffusion problems and produces smooth, high-quality solutions for two-dimensional Poisson and Stokes flows. Its accuracy matches or surpasses advanced approaches such as X-TFC while requiring significantly fewer tunable parameters. In the inverse setting, the same distributional optimization framework accurately identifies unknown coefficients from sparse and noisy sensor measurements, highlighting the robustness and generality of the method.

A key contribution of this work is the extension of KAPI-ELM to \emph{nonlinear} PDEs. Using a curriculum-learning strategy, in which the steady incompressible Navier-Stokes equations are solved through a sequence of linearized problems at increasing Reynolds numbers, KAPI-ELM stably captures flow structures up to $Re=100$. The adaptive kernel distribution evolves coherently with the flow physics, while each curriculum stage preserves the closed-form least-squares character of PI-ELM. These results demonstrate that kernel adaptivity combined with continuation in Reynolds number provides a practical and gradient-free pathway for extending PI-ELM architectures to nonlinear fluid flow.

This study opens several promising avenues for future research. Time-dependent and convection-dominated systems may benefit from transporting kernels along characteristic paths or dynamically updating the distribution in time. Hierarchical Bayesian priors or meta-learning strategies may further automatize the selection of hyperparameter bounds and improve solver robustness. Moreover, the success of the curriculum-based Navier-Stokes simulations suggests that KAPI-ELM can be extended to broader classes of nonlinear multiphysics systems.

Overall, KAPI-ELM offers an interpretable, adaptive, and computationally efficient framework for physics-informed learning. Its combination of analytic training, kernel adaptivity, and low-dimensional optimization makes it a strong candidate for next-generation surrogate models and computational solvers in fluid mechanics, transport phenomena, and general multiphysics applications.

\section*{CRediT authorship contribution statement}
\textbf{Vikas Dwivedi:} Conceptualization, Methodology, Software, Writing - Original Draft, \textbf{Balaji Srinivasan} and \textbf{Monica Sigovan:} Conceptualization and Supervision, \textbf{Bruno Sixou:} Conceptualization, Methodology, Supervision and Writing - Review \& Editing
\section*{Acknowledgments}
This work was supported by the ANR (Agence Nationale de la Recherche), France, through the RAPIDFLOW project (Grant no. ANR-24-CE19-1349-01).

\bibliographystyle{unsrt}  
\bibliography{references}

\end{document}